\documentclass{llncs}
\usepackage{makeidx}  
\usepackage{psfig}
\usepackage{epsf}

\usepackage{times}
\usepackage{amsmath}
\newcommand{\comment}[1]{}
\newcommand{\ignore}[1]{}
\newcommand{\longversion}[1]{}

\def\EQ{\begin{equation}}
\def\EN{\end{equation}}

\def\RR{\hbox{I\kern-.2em\hbox{R}}}

\newcommand{\rem}[1]{}

\newcommand{\norm}[1]{||{#1}||}
\begin{document}

\title{\huge\bf 
\vspace{1cm} 
\Large \bf Recovering Epipolar Geometry from Images of \\Smooth Surfaces}
\author{Oleg Kupervasser\\}
\institute{Industrial Engineering and Management \\The Technion, \\ Haifa 32000, Israel}
\maketitle

\begin{abstract} We present four methods for recovering the epipolar geometry from images of smooth surfaces. In the existing methods for recovering epipolar geometry corresponding feature points are used that cannot be found in such images. The first method is based on finding corresponding characteristic points created by illumination (ICPM - illumination characteristic points method). The second method is based on correspondent tangency points created by tangents from epipoles to outline of smooth bodies (OTPM - outline tangent points method). These two methods are exact and give correct results for real images, because positions of the corresponding illumination characteristic points and corresponding outline are known with small errors. But the second method is limited either to special type of scenes or to restricted camera motion. We also consider two more methods which are termed CCPM (curve characteristic points method, green curves are used for this method on Figures) and CTPM (curve tangent points method, red curves are used for this method on Figures), for searching epipolar geometry for images of smooth bodies based on a set of level curves (isophoto curves) with a constant illumination intensity. The CCPM method is based on searching correspondent points on isophoto curves with the help of correlation of curvatures between these lines. The CTPM method is based on property of the tangential to isophoto curve epipolar line to map into the tangential to correspondent isophoto curves epipolar line. The standard method termed SM (standard method, blue curves are used for this method on Figures) and based on knowledge of pairs of the almost exact correspondent points, has been used for testing of these two methods. The main technical contributions of our CCPM method are following. The first of them consists of bounding the search space for epipole locations. On the face of it, this space is infinite and unbounded. We suggest a method to partition the infinite plane into a finite number of regions. This partition is based on the desired accuracy and maintains properties that yield an efficient search over the infinite plane. The second is an efficient method for finding correspondence between points of two closed isophoto curves and finding homography, mapping between these two isophoto curves. Then this homography is corrected for all possible epipole positions with the help of evaluation function. A finite subset of solution is chosen from the full set given by all possible epipole positions. This subset includes fundamental matrices giving local minimums of evaluating function close to global minimum. Epipoles of this subset lie almost on straight line directed parallel to parallax shift. CTPM method was used to find the best solution from this subset. Our method is applicable to any pair of images of smooth objects taken under perspective projection models, as long as assumption of the constant brightness is taken for granted. The methods have been implemented and tested on pairs of real images. Unfortunately, the last two methods give us only a finite subset of solution that usually includes "good" solution, but doesn't allow us to find this "good" solution among this subset. Exception is the case of epipoles in infinity. The main reason for such result is inaccuracy of assumption of constant brightness for smooth bodies. But outline and illumination characteristic points are not influenced by this inaccuracy. So, the first pair of methods gives exact results.

\end{abstract}


\section{Introduction}

Recovering a three-dimensional shape from a sequence of 2D images has many applications in areas as diverse as autonomous navigation, object recognition, and computer graphics. Solving this problem requires appropriate camera parameters and correspondence between points in different images. Epipolar geometry plays a central role in extracting correspondence between points in different images. For each point in one image epipolar geometry determines a single line, called an epipolar line, in the other image on which its corresponding point is incident.

This paper presents methods for determining the epipolar geometry of a pair of images under weak (epipoles in infinity) and full perspective projection models. We assume uncalibrated images of smooth surfaces. The pair of images may be taken from any two viewpoints distant from each other as long as they satisfy the assumption of constant brightness which presupposes that corresponding points in the different images have the same value of intensity. This assumption holds when the reflectance model of the imaged surface is independent of the viewpoint. We also discuss less general cases such as the weak perspective projection model, calibrated cameras, and a setup similar to the one suggested by \cite{Cross99}, where the images contain a plane whose homography can be computed.

 Epipolar geometry is often represented by the fundamental matrix 
\cite{Faugeras92,Luong96,FaugerasBook} The standard method for recovering the epipolar geometry is that by computing the fundamental matrix from a set of corresponding features in the two images such as points or lines  (e.g., \cite{Hartley95,Torr97,Luong96,LH81,Zhang95,Zhang98}). However, for images of smooth surfaces which we consider in this paper, reliable extraction of image features is often impossible.

The first pair of method gives exact result for real, practical images. The first method is based on finding corresponding characteristic points created by illumination (ICPM). The second method is based on correspondent tangency points created by tangents from epipoles to outline of smooth bodies (OTPM). The second method for recovering the epipolar geometry of smooth objects is based solely on the objects' outline (e.g., \cite{Cipolla95,Cross99,Ponce02}). Such method is limited to either restricted motion or to a relatively rich scene with sufficient number of special points along the objects outline. The second pair of methods is general and independent of the occluding contour and the camera motion. However, in these methods a significantly larger space search must be considered, so it can work only when the assumption of constant brightness is satisfied.

The second pair of methods offered by us termed as CCPM and CTPM is for searching epipolar geometry for images of smooth bodies. The CCPM method is based on searching correspondent points on isophoto curves with the help of correlation of curvatures between these lines. The CTPM method is based on property of the tangential to isophoto curve epipolar line to map into the tangential to a correspondent epipolar line of isophoto curves. The standard method termed SM and based on knowledge of pairs of the almost exact correspondent points was utilized for testing these two methods.

The main technical contributions of our method are following. The first one is an efficient method for finding correspondence between points of two closed isophoto curves and finding homography, mapping between these two isophoto curves. Then this homography is corrected for all possible epipole positions with the help of the evaluation function. A finite subset of solution is chosen from the full set given by all possible epipole position. This subset includes fundamental matrices giving local minimums of evaluating function close to the global minimum. Epipoles of this subset lie almost on straight line directed parallel to parallax shift. CTPM method was used to find the best solution from this subset.

 The next contribution consists of bounding the search space for epipole locations. On the face of it, this space is infinite and unbounded. We suggest a method to partition the infinite plane into a finite number of regions (see Figure 1). The suggested plane partition maintains desired resolution of the system, when possible. In addition, it maintains a probabilistic equal hit measure of epipolar lines. Roughly speaking, probabilistically the size of each subset of epipolar lines we assigned for each region is equal. This property contributes to the efficiency of the search

\begin{figure}[htb]
\centerline{
\begin{tabular}{c}
 \psfig{figure=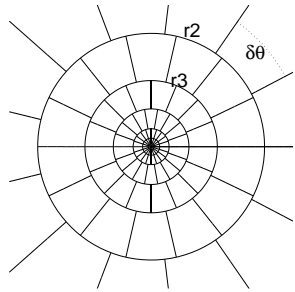,width=1.5 in}
\end{tabular}
} \caption { A  schematic drawing of the infinite plane partitioned into regions.}
\label{fig:all-regions}
\end{figure}

The rest of the paper is organized as follows. We begin by presenting the first pair of method for recovering the epipolar geometry from a pair of uncalibrated images of smooth surfaces (Section~\ref{sec:method0}). The next step is the second pair of methods (Section~\ref{sec:method}). The method is presented for images taken under the perspective projection model. The implementation of these method and the results of running correspondent algorithms on real images are presented in Subsection~\ref{sec:implementation}. Finally we summarize and conclude in Section~\ref{sec:summary}.

\section{The first pair of exact methods.}\label{sec:method0}
\subsection{Method of characteristic points of illumination intensity for smooth bodies (ICPM)}\label{sec:method1}

For searching epipolar geometry corresponding points on two images are used. The assumption of constant brightness was used. This means that illumination intensity of the corresponding points on the both images is identical. Usually the corresponding points are some corner points of bodies. As on smooth bodies such points are lacking, it is usually assumed that this procedure is not applicable. However, it is not absolutely true. Though the corner points on smooth bodies are really lacking, illumination of smooth bodies creates similar characteristic points. For example, even illumination of a spherical body creates a point of maximum of illumination intensity which is easily registered on a pair of the images. We will describe here all characteristic points which illumination creates on smooth bodies.

1) A point of minimum of illumination intensity.

2) A point of maxima of illumination intensity.

3) A saddle point of illumination intensity.

4) A positive cusp (peak) point: a point with a high and positive value of a curvature radius on a level curve of constant illumination intensity.

5) A negative cusp (peak) point: a point with a high and negative value of a curvature radius on a level curve of constant illumination intensity.

6) A curvature point of inflexion: a point on a level curve in which change of the sign of curvature radius occurs. 

The characteristic points types 1-3 are searched as a point with zero derivatives along axes x and y. Curvature in points 4-5 is searched by selection of a series of level lines with some certain step on illumination intensities and searching on these lines of extremes of a curvature radius with a value of curvature radius more than some chosen threshold. For calculation of a curvature radius in points on a level curve the formula \ref{eq:F1} is used. Points of the type 6 are registered by changing of a sign of a curvature radius. After all characteristic points are discovered, it is necessary to discover their correspondence for two images. It can be made by the following methods:

1) Corresponding points should have the same types described above.

2) They should have almost equal illumination intensity.

3) For corresponding points of 1-3 types the two principal curvature radiuses should be close in values. (Curvature radiuses calculated for an illumination intensity surface as functions of x and y in these points).

4) For corresponding points of 4-5 types curvature radii of a level curve should be close in values.

5) For corresponding points the nearest characteristic points along a level curve should correspond to the similar points on the second image and have the same types with them.

6) Comparing local neighborhoods of points, also as it is made in method SIFT \cite{KaFrShTe02,KaFrShTe03,KaFrShTe04}.

After determination of corresponding points their final correspondence and the epipolar geometry is searched by the method RANSAC \cite{ZissermanBook}. But it is better to use the specific version of this method \cite{KaFrShTe05}. It is appropriate for the case when the majority of the discovered points can lie close to a single plane and only small part of these points is considerably out of this plane. It is a frequent case for smooth bodies.

\subsection{Implementation of the first method (ICPM) for real images}\label{sec:method111}

\begin{figure}[htb]
\centerline{
\begin{tabular}{c}
 \psfig{figure=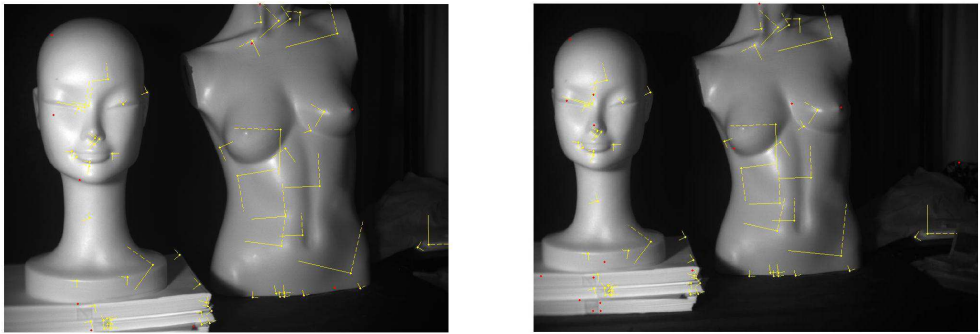,width=6 in}
\end{tabular}
}
\caption {Correspondent points on two images.}
\label{fig:all-regions1}
\end{figure}

Numerical calculation demonstrates that the method SIFT \cite{KaFrShTe02,KaFrShTe03,KaFrShTe04} discovers many of the characteristic points featured above (Figure ~\ref{fig:all-regions1}).

In Figure ~\ref{fig:all-regions1} about 30 pairs of the corresponding points are marked by yellow color. These points are found by SIFT method and part of them lie on smooth bodies (sculptures).
But \emph{all} these points on smooth bodies concern to one of 1-5 types of the characteristic points featured above. Correspondence of the points was found by the 6th method (from the six methods featured above). Numerical calculation shows that these pairs of points give enough information to precisely calculate the epipolar geometry by the usual methods. This result confirms efficiency of the described method. This efficiency is explained by the fact that the characteristic points are found with split-hair accuracy. Indeed, intensity of illumination along these bodes varies slowly and smoothly, except for the cases of characteristic points featured above. It results in small errors for the epipolar geometry. 

\subsection{Method of tangents to an outline of smooth bodies (OTPM)}\label{sec:method2}

However, there is one more method that has also split-hair accuracy - a method of tangents to the outline of smooth bodies. (This method is discoursed in details in \cite{Cipolla95,Cross99,Ponce02}.) Let A be the tangency point of a tangent to a smooth bodies outline from the epipole on the first image. And B is the tangency point of the correspondent tangent to the correspondent outline of smooth bodies from the second epipole on the second image. It is proved in \cite{Ponce02}, \cite{ZissermanBook} that these two points are correspondent points of the images. Suppose that we have many smooth bodies and, consequently, many outlines of smooth bodies. Then, by dividing image planes on meshes we consider all possible methods for putting epipoles to these meshes. It is possible to find pair epipoles giving the minimal error for the described property of tangents. The described method is very similar to a method of tangents to level curves of illumination intensity described below (CTPM). However, the method described here is much more exact than the method described below. Really, level curves of illumination intensity are found insufficiently precise both because of inexactness of assumption of constant brightness and because of noise. This noise leads to major errors in definition of corresponding points, especially for smooth bodes. Indeed, intensity of illumination along these bodes varies slowly and smoothly, except for the cases of characteristic points featured above, and outlines of smooth bodies. This fact results in the split-hair accuracy of the described here two methods with respect to the two methods described below. In case of a small number of smooth bodies, the described here method of tangents can be used only for validation and improvement of the method of characteristic points or for only finding a \emph{set} of possible epipolar geometries.

The second pair of two methods described below (CCPM and CTPM; these methods use level curves of constant illumination intensity) find corresponding points with much bigger errors. It occurs because assumption of constant brightness is not exact. That results in serious errors during finding epipolar geometry. Therefore, these methods can be used only for validation and improvement of the present method, or only for finding a \emph{set} of possible epipolar geometries.

\section{Two methods using level curves of illumination intensity} \label{sec:method}

 In this section we present our method for recovering the epipolar geometry of two images, by determining the fundamental matrix $\cal F$. (A pair of corresponding points in the two images, ${\bf x}$ and ${\bf x'}$, must satisfy to ${\bf x}{\cal F}{\bf x'}^T=0$.) The purpose of methods stated in this section is reconstruction of epipolar geometry of two 2D images of 3D smooth bodies photographed from various positions. For finding epipolar geometry in a case of non-smooth bodies we use correspondence of the reference points, such as corners, for example. In case of smooth bodies these corner points are primely not present which makes finding correspondence quite a problem to solve. Nevertheless, it can be tried to be solved.  A basis for this purpose is assumption of constant brightness. It is supposed that corresponding points in different images have one and the same value of intensity. This assumption is proved for the following conditions \cite{ZissermanBook}:

1) The Lambert law of reflections of surfaces. 

2) An invariance of position and intensity of irradiating sources and also objects
position invariance during shooting. 

3) A small angle between an optical axis and a direction on any of object point, i.e.
the object should be near to the optical axis and have small angular sizes.

4) Constant interior camera calibration for both images.

It is necessary to note that because of infraction of these requirements and also because of presence of noise the assumption of constant brightness is carried out very approximately. It creates all further problems. The curve wherein all points have identical intensity is called an isophoto curve. Correspondence of reference points as was in the case of non-smooth bodies, is exchanged by correspondence isophoto curves for smooth bodies. We shall see further that the same inexactness of assumption of constant  brightness leads to similar problems in the next methods too

\subsection{ Description of the two methods for finding correspondence and reconstruction of epipolar geometries.}
\label{sec:subsets}

     The first method presented by us allows passing from correspondence isophoto curves to correspondence of points on these lines. We shall call this method CCPM. For  finding this correspondence we shall use closed isophoto curves. Generally lines on object of shooting which image are isophoto curves, they are complicated space curves. However, in a large number of practical cases all points of such curve almost lie in the same plane and, accordingly, connection between isophoto curves on various images is very close to homography. In this case homography cusp transfers in cusp, and the inflection point transfers in an inflection point. Accordingly, for the composite curve with strongly varying curvature there is a correlation between quantities of curvature in the correspondent points. Moreover, in the majority of practical cases homography between isophoto curves is very close to the similarity of transformation including translation, rotation in plane and scaling. Curvature of curves in the correspondent points is identical up to constant for similarity transformation. For finding isophoto curves, curvature in any point is not necessary to be searched for the shape of this line. There is a simple formula allowing finding curvature on the basis of intensity distribution in a neighborhood of this point:

 \begin{eqnarray}
   k=-\mbox{div} \left( \frac{\bigtriangledown \phi}{|\bigtriangledown \phi|}    \right) \label{eq:F1}
\end{eqnarray}
 , where $k$ is curvature of isophoto curve and $\phi$ is intensity.

     Assuming that correspondence between isophoto curves is close to similarity transformation, choosing complicate closed isophoto curves with strongly changing curvature k and carrying out rescaling by multiplying curvature on length of the curve (the formula \ref{eq:F2}, \ref{eq:F3}) we can find correspondence between these curves.

\begin{equation}
k^r=k \times L \label{eq:F2}
\end{equation}

A mean value of rescaled curvature $k^r$ doesn't depend on length $L$ of the curve 

\begin{equation}
k^r_{\mbox{mean}}=\frac{1}{L}  \int_0^L k^r\,dl=\frac{1}{L}
\int_0^L k \times L \,dl=  \int_0^L k \,dl = \int_0^{2\pi}
d\varphi=2\pi\label{eq:F3}
\end{equation}

     It can be achieved by finding correlation between two functions: the rescaled curvature $ k \times L$ as functions of the rescaled distance $\frac{r}{L}$  from the some point on some chosen isophoto curves, and the similar function for the correspondent isophoto curve where the rescaled distances are calculated from a some chosen correspondent point on this correspondent isophoto curve. Some correspondence between these two points which gives the maximum of correlation of these two functions is considered as correct correspondence. Other correspondent points are on the identical rescaled distances from the already found pair of correspondent points. That is the property of similarity transformation. It is also necessary to note that all the above-mentioned concerns to the smoothed image and functions; otherwise the noise disturbs all correlations. I.e. it is required to use filtration of images before searching isophoto curves and filtration of functions for the rescaled curvature for searching maximum of correlation.

 For testing of correspondence instead of function of the rescaled curvature in current point, the rescaled distance from a centre of gravity for current point can be used. Such function should give a maximum of correlation approximately under the same requirements as the function of the rescaled curvature. Choosing various pairs of correspondent isophoto curves, we get a set of correspondent points. The problems is that for any alone smooth body all these points lie in very close to almost parallel planes. These planes lie on almost identical distance from the camera, i.e. the parallax will be very small, and the fundamental matrix determining epipolar geometry will be defined ambiguously. To solve this problem, we chose two smooth bodies located on various distance from the camera, and the parallax is large enough for isophoto curves on these bodies. It is necessary to note, that this correspondence between points is found on the basis of assumption about similarity and, hence, is not exact. These errors are added to the errors caused by inexactness of assumption of constant brightness. Therefore, the fundamental matrix received by conventional methods on the basis of these pairs of the correspondent points is far from its true value. For finding its more precise value it is necessary to involve additional reasons. First of them is connected with the fact that in most cases points of a 3D-prototype isophoto curves usually lie almost in the same plane. Accordingly, connection between isophoto curves, as we already noted above, is close to homography. Knowing pairs of the correspondent points, we can find $3 \times 3$ matrix homography $H$, linking these pairs of points. For this purpose a normalized DLT algorithm is used \cite{ZissermanBook}. The result can be improved by minimization of the function determined as the sum of absolute values of distances between homographic mappings of isophoto curve points to nearest points on correspondent isophoto curve. Then this sum is added to the sum, received by the same way but two images are interchanged, and mean value of these errors is calculated:

\begin{eqnarray}
S=\left. \left(\sum_{i=1}^N \left(\min_j d({\bf x}^\prime_j,H{\bf
x}_i)+ \min_j
d({\bf x}_j,H^{-1}{\bf x}_i^\prime)\right)\right)\right/ 2N  \nonumber\\
\label{eq:F4}\\
{\mbox{, where  }}d({\bf a},{\bf
b})=((a_1/a_3-b_1/b_3)^2+(a_2/a_3-b_2/b_3)^2)^{1/2}\nonumber
\end{eqnarray}

Where $d$ is distance between points with homogeneous coordinates, ${\bf x}_i$, ${\bf x^\prime}_i$ are homogeneous vectors of correspondent points on isophoto curves.  Points ${\bf x}_i$ are on the first image, points ${\bf x}_i^\prime$ are on the second image. N - number of pair of correspondent points, $H$- evaluated homography between closed isophoto curves. Making value of a corner element of matrix $H(3,3)$ equal to unity, we search for the local minimum of function defined by formula \ref{eq:F4} with respect to other elements of this matrix. For the initial iteration the value of matrix $H$ will be used which is already found above by normalized DLT algorithm. For search of the local minimum we use standard function of MATLAB package ${\bf fminsearch()}$.The received minimal value of function defined by formula \ref{eq:F4}
, we shall denote as $Smin$. It is necessary to note that resulting homography is found by non-exact methods and, accordingly, also is not exact.

Suppose that we know homogeneous coordinates of epipole on the second image. Using also the homography found above, we can calculate a fundamental matrix from the following formula:

\begin{eqnarray}
   F={\bf [e^\prime]_\times}H\nonumber\\
    \label{eq:F5}\\
\mbox{where  }{\bf [e^\prime]_\times}= \left[\begin{array}{ccc}
  0       & -e'_3            & e'_2 \cr
e'_3       & 0           & -e'_1 \cr
 -e'_2      &e'_1      & 0
\end{array}\right]\nonumber
 \end{eqnarray}

Where ${\bf e}^\prime$ - is a homogeneous vector of epipole on the second image, H is some homography between two images. For search of candidates in epipole we divide the plane into a finite number of regions as it is described in the second part of this section. In each region we select just one epipole. Let's choose a function for evaluation of "quality" of a fundamental matrix. We will map each point on the first image to an epipolar line in the second image. Then we shall calculate the absolute value of distance from the correspondent point on the second image to this epipolar lines. We shall take the sum of all these distances, and then we shall add this sum to the similar sum received by the same way but with interchange of images. Then we shall find the mean value for components of these sums.

\begin{eqnarray}
Z=\left. \left(\sum_{i=1}^N \left( d_l({\bf x}^\prime_i,F{\bf
x}_i)+
d_l({\bf x}_i,F^{T}{\bf x}_i^\prime)\right)\right)\right/ 2N  \nonumber\\
\label{eq:F6}\\
{\mbox{, where  }}d_l({\bf x},{\bf l})=(\frac{|{\bf x^T}{\bf l
}|}{x_3\sqrt{l_1^2+l_2^2}})\nonumber
\end{eqnarray}

Where $d_l$ distance between a point with homogeneous coordinates and an epipolar line,${\bf x}_i$, ${\bf x^\prime}_i$ are homogeneous vectors of correspondent points on isophoto curves.
Points ${\bf x}_i$ are on the first image, points ${\bf x}_i^\prime$ are on the second image. N - number of pair of correspondent points, $F$- evaluated fundamental matrix. Choosing some candidate in epipole and having taken as initial value for homography matrix H found as described above by minimization with help formula \ref{eq:F4},  we can carry out further iterations for and find a local minimum of the function described by the formula \ref{eq:F6}.

    Found homography will be more precise then its initial value, because initial homography was only the first approximation. In order not to "run" too far from initial homography, we shall assume that deviation S determined by the formula \ref{eq:F4} cannot be more then  $1.5Smin$ where $Smin$ is its minimal value found above. It can be achieved by introduction of the additional term in the formula \ref{eq:F6} equal to zero for $S <1.5Smin$ and fast increasing for $S> 1.5 Smin$. Search of a local minimum can be carried out with use of the above mentioned function of MATLAB package ${\bf fminsearch()}$.For each candidate in epipole with the number i we find optimum fundamental matrix $Fi$. The best of them can be defined as giving a minimum of function \ref{eq:F6}. We shall term this minimum value as $Zmin$.

    Unfortunately, since the found pairs of the correspondent points are only approximate, the minimum of function 6 can not give the true value of a fundamental matrix. To improve the result we can take not alone fundamental matrix relevant to $Zmin$, but all local minimums of function $Z$ ($Fi$) from argument i (epipole correspondent to i=1 is in the center of image and run away from center to maximal allowed distance along spiral curve for larger $i$ (yellow spiral on figure \ref{fig:fige1} \texttt{\textbf{(h)}}) ) , where i is a serial number of epipole. From all found local minimums we choose only that $Z <1.34Zmin$, where Z is calculated from formula \ref{eq:F6}.

      It will give us not hust one solution but a large set of candidates for the solution. For finding of the best of them the second reason and the second method of search of a fundamental matrix which we shall term CTPM can be used. It is based on the following property of epipolar geometry that is correct for assumption of constant brightness: epipolar line tangential to isophoto curves maps to epipolar line, is also tangential to correspondent isophoto curves. To evaluate how precisely this property carried out, we can introduce the following function:

\begin{eqnarray}
Z=\left. \left( \sum_{i=1}^{N_1} \min_{j \in M_i^\prime} d_l({\bf
y}^{\prime \mbox{tangent}}_j,F{\bf x}_i)+
\sum_{i=1}^{N_2} \min_{j \in M_i} d_l({\bf y}_j^{\mbox{tangent}},F^{T}{\bf x}_i^\prime)\right)\right/ (N_1+N_2)  \nonumber\\
\label{eq:F7}\\
{\mbox{, where  }}d_l({\bf x},{\bf l})=(\frac{|{\bf x^T}{\bf l
}|}{x_3\sqrt{l_1^2+l_2^2}})\nonumber
\end{eqnarray}

\begin{figure}[htb]
\centerline{ \psfig{figure=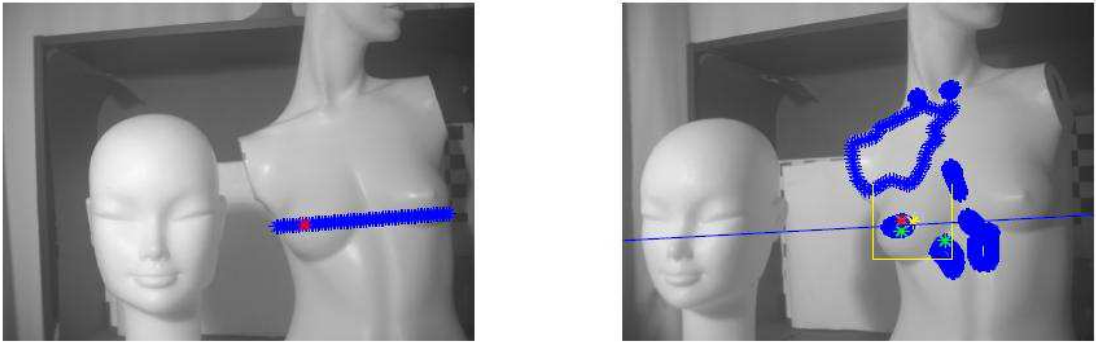,width=6in}}
\caption{Finding correspondent tangential point.}
\label{fig:figol}
\end{figure}

Where $d_l$ distance between a point with homogeneous coordinates and an epipolar line, ${\bf x}_i$, ${\bf x^\prime}_i$ are homogeneous vectors of globalmaximum points of intensity on epipolar lines. Points ${\bf x}_i$ are on the first image, points  ${\bf x}_i^\prime$ are on the second image. $N_1$ - number of points on first image that have correspondent tangent points  $y_j^{\prime \mbox{tangent}}$ on the second image with respect to condition described below. $N_2$ - number of points on second image that have correspondent tangent points $y_j^{\mbox{tangent}}$  on the first image with respect to condition described below. $F$- evaluated fundamental matrix. Let's choose a certain set of epipolar lines  on the first image transiting through a smooth body.We shall consider intensity along pieces of epipolar lines lying inside the image of a smooth body. Points of extremum give us a point of tangency epipolar line with the isophoto curve. We shall choose some epipolar line, and among points of extremum we shall choose a global maximum. We shall find the correspondent epipolar line on the second image. Also we shall find all isophoto curves on the second image having the same quantity of intensity, as in the point of a global maximum found above on the first image. Let's carry out all possible tangential to these isophoto lines from epipole on the second image. Among all points of a tangency we shall choose what is closest to epipolar line and we shall find its distance to the epipolar line. We shall make the same calculations for all epipolar lines, and then we shall calculate the sum of all found distances. Then we shall find the similar sum, by images interchanging. Then we shall add together these sums and then we shall calculate the mean value of distance. Thus, we get evaluation for precision of performance of the tangential epipolar lines property described above. However, the algorithm as described above works with big outliers. Actually, among all points of a tangency found on the second image should there is a point correspondent to a global maximum on the first image. Its distance to the correspondent epipolar line gives a current error of a method. Other points of tangency can improve only the result and reduce an error if one of them lies closer to epipolar lines, than this true point of tangency. However, the situation is much more difficult. The true point of tangency can in general miss on the second image. It can be closed by others parts of smooth bodies or can be not found because of an error in assumption of constant brightness. In this situation a point of tangency the closest to epipolar lines can lie very far from this epipolar line because it has no relation to the true tangency correspondent point. Such tangency point consequently gives outlier to evaluated function. To prevent such situation, we offer the following method. We shall take outline curves on the first and second image. We shall find homography, mapping these curves by the same method which we found homography, mapping isophoto curves. We shall find with help of this homography on the second image mapping of a global maximum point, defined above and we shall consider a small circular neighborhood around this mapping (  $3$  percents from the area occupied with a smooth body projection on the second image)(figure \ref{fig:figol}).

       The true point of tangency should lie in this neighborhood. Therefore, we shall view only the points of a tangency lying in this neighborhood. If any point of tangency in this neighborhood is not present, we throw out this case from our consideration. This method will prevent appearance of outliers in evaluated function. The similar method can be used for search correspondent isophoto curves. Actually, if we have one isophoto curve on the first image, it usually corresponds to not alone but several isophoto curves on the second image. Which one of them to choose? The one which is closest to a homographic mapping of the first image isophoto curve. We use homography received from outline curves. We can use the above described CTPM evaluation method that excludes outliers to evaluate all possible fundamental matrixes and choose the one what gives the minimal error. However, it is a too long way. For testing by the second CTPM method we shall choose only those fundamental matrixes which have passed the filter of the first CCPM method. Thus, we get a final evaluation of fundamental matrixes. Precision of this evaluation is restricted only to precision of assumption of constant brightness.

\subsection{Searching for  epipoles}
\label{sec:partition}

In this section we present a partition of the infinite plane into a finite number of regions. All points in a given region will be treated, under this partition, as a single epipole. Therefore, all lines intersecting a given region will be members of its epipolar line set. As we explained below, the size of each region is proportional to its location with respect to the image.

Our plane partition is defined by concentric circles with a set of radii $\{r_i\}_{i=1}^k$, where $r_i\geq r_{i+1}$ for each $1\leq i\leq k$ and $r_1=\infty$. The center of these circles is the image
center. Let each region be defined by the 4-tuple, $(\theta,\delta\theta_i,r_i,r_{i+1})$, where $\theta_i$ and $\theta_i+\delta\theta_i$ are two angles, and $r_i$ and $r_{i+1}$ are two radii (see Figure~\ref{fig:all-regions}).  The parameters of this partition are the set of $r_i's$ and the set of
$\delta\theta_i's$. For a given ring, there are two degrees of freedom to define the set of regions in that ring: the ring length,  $r_i-r_{i+1}$ and the region width, $\delta\theta_i$. These parameters are set so that the partition maintains the system resolution and the equal hit measure. Later we define these two properties. For simplicity, we assume a circular image with radius of one.
	
\vspace{-0.4cm}
\subsubsection*{System resolution:}

Any vision system is limited by the accuracy of the measurements. We define {\it system resolution} to be $\gamma$   if it does not discriminate between two image lines passing through a point when the difference in the line directions is $\gamma_0 \leq \gamma$.

\begin{figure}[htb]
\centerline{ \psfig{figure=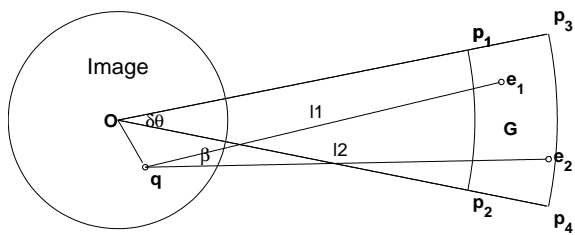,width=3in}}
\caption{$\vec q$ is an image point, $\vec e_1$ and $\vec e_2$ are two points in a given region $G$.  The angle $\beta=acos((\vec{e_1}-\vec q)(\vec{e_2}-\vec q)^T/\norm{(\vec
{e_1}-\vec q)}\norm{\vec {e_2}-\vec q})$. If $\beta <\gamma$ where $\gamma$ is the system resolution, then the system cannot discriminate between the epipoles $\vec {e_1}$ and $\vec {e_2}$.}
\label{fig:resolution}
\end{figure}

We say that the partition maintains the system resolution when the system cannot discriminate between two candidate epipoles which are located in the same region. Formally, Assume the system resolution is . Let $G$ be a given region and $\vec e_1,\vec e_2\in G$ be two points (see Figure ~\ref{fig:resolution} ). Let $l_1$ and $l_2$ be two lines connecting an image point $\vec q$ and the two points $\vec e_1$ and $\vec e_2$, respectively. When the angle between $l_1$ and $l_2$ is less than the system resolution $\gamma$,  the system cannot discriminate between these lines. More generally, for a given region $G$ we define $\alpha_G(\vec q)$ to be the maximal angle between the image point $\vec q$  and any two points in the region. Let $\alpha_G=\max\limits_{\vec q\in Image}{\alpha_G(\vec q)}$ then the system resolution is maintained for G.

Note that the system resolution cannot be maintained in regions which overlap the image or are very close to the image. The closer the point q is to the region $G$ the larger $\alpha_G(\vec q)$  becomes. In particular when $\vec q\in G$ then $\alpha_G(\vec q)=2\pi$.

\vspace{-0.4cm}
\subsubsection*{Hit measure:}

Roughly speaking, the equal hit measure property guarantees that the number of epipolar lines considered for each region is probabilisticly equal. We define the hit measure of a region, $HM(G)$, to be the probability of a random epipolar line intersecting a region $G$. Each candidate epipolar line we consider is defined by an image point and a single direction. We assume that these epipolar lines are generated by a uniform distribution of points in the image and uniform distribution of directions. Formally, the following integral computes $HM(G)$.

\begin{equation}
HM(G)={1\over 2\pi}\iint\limits_{x^2+y^2\leq 1} \alpha_G(x,y) dx
dy
\end{equation}

By changing the integral variables from $x$ and $y$ to the polar coordinates of the image points, $\phi$ and $r$, we obtain:

\begin{equation}
HM(G)={1\over 2\pi}\int_{0}^1\int_{0}^{2\pi} r
\alpha_G(r\cos(\phi),r\sin(\phi)) d\phi dr \label{eq:VP2}
\end{equation}

($r$ is the determinant of the Jacobian of the exchange variables).

In order to have a probabilistically equal number of points in each region, we would like $HM(G)$  to be identical for all regions.

\vspace{1cm} In the Appendix we describe how to set the partition parameters, $r_i$ and $\delta\theta_i$, in order to maintain the above conditions. We distinguish between three types of regions which we analyze separately. They include infinite regions on the outermost ring, regions within the image and close to the image such that the system resolution condition cannot be maintained and intermediate regions. The parameters depend on the desired resolution of the system, $\gamma$.

\subsection{Functions for testing of correspondence.}
\label{sec:eval}

\begin{figure}[htb]
\centerline{ \psfig{figure=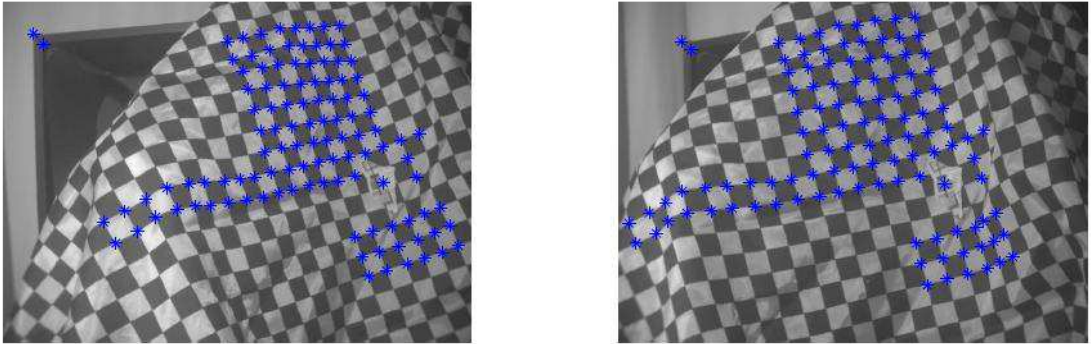,width=6 in}}
\caption { Covering smooth bodies with a fabric with a texture of a chess board.} \label{fig:figol1}
\end{figure}

Let's assume that we have some fundamental matrix. We shall introduce three ways of an evaluation as far as it is close to the true fundamental matrix. The first evaluation shall be termed CCPM. It is based on a set of the correspondent points found from correlation of curvatures for isophoto curves and given by the formula \ref{eq:F6}. The second evaluation shall be termed CTPM by us. It is based on property of the tangential to isophoto curves epipolar lines to map also into the tangential to correspondent isophoto curves epipolar line and is given by the formula \ref{eq:F7}. The third evaluation shall be termed SM. It is based on a set of almost precise correspondent points received by covering smooth bodies with a fabric with a texture of a chess board (figure \ref{fig:figol1}). Then we photograph pair of images of this fabric from the same points of shooting from which images of smooth bodies have been received also. Evaluated function is given by the following formula:

\begin{eqnarray}
B=\left. \left(\sum_{i=1}^N \left( d_l({\bf z}^\prime_i,F{\bf
z}_i)+
d_l({\bf z}_i,F^{T}{\bf z}_i^\prime)\right)\right)\right/ 2N  \nonumber\\
\label{eq:F8}\\
{\mbox{, where  }}d_l({\bf x},{\bf l})=(\frac{|{\bf x^T}{\bf l
}|}{x_3\sqrt{l_1^2+l_2^2}})\nonumber
\end{eqnarray}

Where $d_l$    is distance between a point with homogeneous coordinates and an epipolar line, ${\bf z}_i$, ${\bf z^\prime}_i$ are homogeneous vectors of correspondent points on fabric with a texture of a chess board. Points ${\bf z}_i$  are on the first image, points ${\bf z}_i^\prime$ are on the second image. N - number of pair of correspondent points, $F$-  evaluated fundamental matrix. From this almost precise set of the correspondent points by a method of the normalized 8-dot algorithm (for number of points $n>8$) \cite{Cross99}.  almost precise fundamental matrix can be received. We shall term this third standard method for fundamental matrix finding as SM.

\subsection{Implementation} \label{sec:implementation} 

For testing of algorithms nine pairs of images of a pair of smooth bodies have been made. A female head and a female bust were used as smooth bodies. Accordingly, we have nine various cases of algorithms implementation. Images differ with an arrangement of the camera position or an arrangement of smooth bodies. In cases 3,4, and 6 two subcases are chosen, distinguished by a choice of isophoto curves on the basis of which calculation of the correspondent points by a CCPM method was made. As result of use of this method we get a finite set of candidates for a role of a fundamental matrix as it has been described earlier in Section ~\ref{sec:method}. For each of cases or subcase, eight figures are given:

1) In the first figure we can see a set of almost precisely correspondent points and a set of correspondent to them epipolar lines of a fundamental matrix received by almost precise SM method.

2) In the second figure we can see a set of almost precisely correspondent points and a set of correspondent to them epipolar lines of a fundamental matrix. This matrix is one component of the complete set of candidates on the fundamental matrix, received by a CCPM method. SM evaluation gives a minimum for this matrix among all matrices of this set.

3) In the third figure we can see a set of almost precisely correspondent points and a set of correspondent to them epipolar lines of a fundamental matrix. This matrix is one component of the complete set of candidates on the fundamental matrix, received by a CTPM method. CTPM evaluation gives a minimum for this matrix among all matrices of this set.

4) In the fourth figure we can see a set of almost precisely correspondent points and a set of correspondent to them epipolar lines of a fundamental matrix. This matrix is one component of the complete set of candidates on the fundamental matrix, received by a CCPM method. CCPM evaluation gives a minimum for this matrix among all matrices of this set.

5) In the fifth figure we can see a set of the correspondent points lying on correspondent
isophoto curves and received by a CCPM method.

6) In the sixth figure we can see CCPM evaluation for a fundamental matrix as function of the epipole number. This epipole located on the second image. All epipoles lie in regions; the method of division into these regions is described in the Section ~\ref{sec:method}. The first epipole lies at center of a figure and with increase of the number moves from center along spiral figured by yellow color in the eighth figure. Fundamental matrixes are received by a CCPM method. Red asterisks designate the set of candidates giving the best CCPM evaluation.

7) In the seventh figure we can see CTPM, CCPM and SM evaluation for a set of the candidates marked by red asterisks in the sixth figure. The asterisks lying on curves, mark points of a minimum. The asterisks not lying on curves and located after graphics mark CTPM, CCPM and SM evaluation for almost precise fundamental matrix. This matrix is received by a SM method. 

8) In the eighth figure by red daggers are marked centre and edges of the second image. The yellow spiral connects regions in which we place epipoles. Epipoles are located along this spiral, since center of the image. By red asterisk it is designated epipole on the second image of almost precise fundamental matrix received by a SM method. Red, green and blue circles are designated epipoles, correspondent to the fundamental matrixes minimizing a CTPM, CCPM and SM evaluation function among a set of candidates, received by a CCPM method and marked by red asterisks in the sixth figure.

\texttt{\textbf{Case1.}}

Case1 includes the following figures: figure \ref{fig:fige1} \texttt{\textbf{(a)}}\texttt{\textbf{(b)}}\texttt{\textbf{(c)}}\texttt{\textbf{(d)}}\texttt{\textbf{(e)}}\texttt{\textbf{(f)}}\texttt{\textbf{(g)}}\texttt{\textbf{(h)}}.

\texttt{\textbf{Case2.}}

Case2 includes the following figures: figure \ref{fig:fige2}
 \texttt{\textbf{(a)}}\texttt{\textbf{(b)}}\texttt{\textbf{(c)}}\texttt{\textbf{(d)}}\texttt{\textbf{(e)}}\texttt{\textbf{(f)}}\texttt{\textbf{(g)}}\texttt{\textbf{(h)}}.

\texttt{\textbf{Case3a.}}

Case3a includes the following figures: figure \ref{fig:fige3a}
 \texttt{\textbf{(a)}}\texttt{\textbf{(b)}}\texttt{\textbf{(c)}}\texttt{\textbf{(d)}}\texttt{\textbf{(e)}}\texttt{\textbf{(f)}}\texttt{\textbf{(g)}}\texttt{\textbf{(h)}}.

\texttt{\textbf{Case3b.}}

Case3b includes the following figures: figure \ref{fig:fige3b}
 \texttt{\textbf{(a)}}\texttt{\textbf{(b)}}\texttt{\textbf{(c)}}\texttt{\textbf{(d)}}\texttt{\textbf{(e)}}\texttt{\textbf{(f)}}\texttt{\textbf{(g)}}\texttt{\textbf{(h)}}.

\texttt{\textbf{Case4a.}}

Case4a includes the following figures: figure \ref{fig:fige4a}
 \texttt{\textbf{(a)}}\texttt{\textbf{(b)}}\texttt{\textbf{(c)}}\texttt{\textbf{(d)}}\texttt{\textbf{(e)}}\texttt{\textbf{(f)}}\texttt{\textbf{(g)}}\texttt{\textbf{(h)}}.

\texttt{\textbf{ Case4b.}}

Case4b includes the following figures: figure \ref{fig:fige4b}
 \texttt{\textbf{(a)}}\texttt{\textbf{(b)}}\texttt{\textbf{(c)}}\texttt{\textbf{(d)}}\texttt{\textbf{(e)}}\texttt{\textbf{(f)}}\texttt{\textbf{(g)}}\texttt{\textbf{(h)}}.

\texttt{\textbf{Case5.}}

Case5 includes the following figures: figure \ref{fig:fige5}
 \texttt{\textbf{(a)}}\texttt{\textbf{(b)}}\texttt{\textbf{(c)}}\texttt{\textbf{(d)}}\texttt{\textbf{(e)}}\texttt{\textbf{(f)}}\texttt{\textbf{(g)}}\texttt{\textbf{(h)}}.

  \texttt{\textbf{Case6a.}}

Case6a includes the following figures: figure \ref{fig:fige6a}
 \texttt{\textbf{(a)}}\texttt{\textbf{(b)}}\texttt{\textbf{(c)}}\texttt{\textbf{(d)}}\texttt{\textbf{(e)}}\texttt{\textbf{(f)}}\texttt{\textbf{(g)}}\texttt{\textbf{(h)}}.

\texttt{\textbf{Case6b.}}

 Case6b includes the following figures: figure \ref{fig:fige6b}
 \texttt{\textbf{(a)}}\texttt{\textbf{(b)}}\texttt{\textbf{(c)}}\texttt{\textbf{(d)}}\texttt{\textbf{(e)}}\texttt{\textbf{(f)}}\texttt{\textbf{(g)}}\texttt{\textbf{(h)}}.

\texttt{\textbf{Case7.}}

Case7 includes the following figures: figure \ref{fig:fige7}
 \texttt{\textbf{(a)}}\texttt{\textbf{(b)}}\texttt{\textbf{(c)}}\texttt{\textbf{(d)}}\texttt{\textbf{(e)}}\texttt{\textbf{(f)}}\texttt{\textbf{(g)}}\texttt{\textbf{(h)}}.

  \texttt{\textbf{Case8.}}

Case8 include follow figures: figure
 \ref{fig:fige8}\texttt{\textbf{(a)}}\texttt{\textbf{(b)}}\texttt{\textbf{(c)}}\texttt{\textbf{(d)}}\texttt{\textbf{(e)}}\texttt{\textbf{(f)}}\texttt{\textbf{(g)}}\texttt{\textbf{(h)}}.

  \texttt{\textbf{Case9.}}

Case9 includes the following figures: figure \ref{fig:fige9}
 \texttt{\textbf{(a)}}\texttt{\textbf{(b)}}\texttt{\textbf{(c)}}\texttt{\textbf{(d)}}\texttt{\textbf{(e)}}\texttt{\textbf{(f)}}\texttt{\textbf{(g)}}\texttt{\textbf{(h)}}.

For these figures:

\texttt{\textbf{(a)}} the set of exact correspondent points (almost exact set found from the chesslike fabric) and epipolar lines found from a fundamental matrix, received by the SM from these points 

 \texttt{\textbf{(b)}} the set of exact correspondent points and epipolar lines found from a fundamental matrix. This matrix gives a minimum error for set of epipoles candidates defined on figure "f" (Red asterisks) and for the exact set of correspondent points.

 \texttt{\textbf{(c)}} the set of exact correspondent points and epipolar lines found from a fundamental matrix. This matrix gives a minimum error for set of epipoles candidates defined on figure "f" (Red asterisks) and for the correspondent points set defined by the CTPM and current epipoles.

 \texttt{\textbf{(d)}} the set of correspondent points (almost exact set) and epipolar lines found from a fundamental matrix. This matrix gives a minimum error for set of epipoles candidates defined on figure "f" (Red asterisks) and for the correspondent points set defined by the CCPM.

 \texttt{\textbf{(e)}} the set of the correspondent points lying on correspondent isophoto curves and received by the CCPM.

 \texttt{\textbf{(f)}} the error of a fundamental matrix as function of the epipoles set index. These
epipoles located on the second image. All epipoles lie in regions; the method of division into these regions is described in the Subsection~\ref{sec:partition}
. The first epipole lies at center of a figure and with increase of the index moves from center along spiral figured by yellow color in the figure "h". Fundamental matrixes are received form the CCPM correspondent points sets and current epipoles positions. Red asterisks designate the set of candidates giving the minimal local errors.

\texttt{\textbf{(g)}} errors of fundamental matrices as a function of index in subset of the candidates to epipoles marked by red asterisks in figure "f". The fundamental matrices are found from current candidates to epipoles and correspondent points sets(set defined by the CTPM and current epipoles(red line), set defined by CCPM (green line) and exact set (blue line)). Errors are calculated with respect to the same relevant correspondent points sets. Asterisks on the lines mark minimums. The three asterisks (not lying on curves and located after graphics) mark error of the exact fundamental matrix. The exact fundamental matrix is found by SM from the exact set of correspondent points. Errors are found for sets of correspondent points (set defined by the CTPM and exact epipoles (red asterisk), set defined by CCPM (green asterisk) and exact set (blue asterisk)).

 \texttt{\textbf{(h)}} marked by red daggers are the center and edges of the second image. The yellow spiral connects regions in which we place epipoles. Epipoles are located along this spiral, from center of the image. Designated by red asterisk is the epipole on the second image from exact fundamental matrix received by the SM. Rosy points mark the set of epipoles candidates. This set is received by the CCPM and marked by red asterisks in figure "f". Circles are designated epipoles, correspondent to the fundamental matrixes, witch are correspondent to minimum errors (between all rosy points) for CTPM (red circle), CCPM (green circle) and exact (blue circle) correspondent points sets.

From the given figures it is possible to draw the following conclusions:

1) The CCPM method gives the full set of candidates on the true solution. Epipoles of these matrixes lie approximately on straight line. What is this straight line? We shall take homography (relevant to evaluation Smin in a CCPM method) between two isophoto curves, lying on a sculpture of the female head on two images. We use this homography to find on the second image mapping of isophoto curve belonging to a bust on the first image. This mapping is shifted with respect to correspondent isophoto curve belonging to a bust on the second image, because of a parallax. The direction of this shift gives us the required straight line.

2) Almost always (in 10 cases from 12, and besides two "bed" subcases actually corresponding to the same pair of pictures) among set of solutions found by CCPM method there is a solution which is very close to the almost precise solution received by a SM method. This solution gives minimum of a SM evaluation for this set.

3) The solutions correspondent to a minimum of a CTPM or CCPM evaluations for this set, less than in 50 percents of cases are close to almost precise minimum (in 5 cases from 12 for CTPM and in 3 cases from 12 for a CCPM evaluation).

4) CTPM or CCPM evaluations for almost precise solution, received by the SM method are not in minimum while SM evaluation always corresponds to a minimum. These three (CTPM, CCPM and SM) evaluations for SM method solution are compared to similar evaluations for the set solution received by CCPM method.

\section{Summary and discussion}
\label{sec:summary}
 
We present four methods for recovering the epipolar geometry from images of smooth surfaces. The existing methods for recovering epipolar geometry use corresponding feature points that cannot be found in such images. The first method is based on finding corresponding characteristic points created by illumination (ICPM – illumination characteristic points' method). The second method is based on correspondent tangency points created by tangents from epipoles to outline of smooth bodies (OTPM – outline tangent points' method). These two methods are exact and give correct results for real images, as positions of the corresponding illumination characteristic points and corresponding outline are known with small errors. But the second method is limited either to special type of scenes or to restricted camera motion.

The level curves of constant illumination intensity (used for the second pair of methods) are found precise but insufficiently both because of inexactitude of constant brightness assumption and because of noise. This noise leads to major errors in definition of corresponding points found from the level curves especially for smooth bodes. Indeed, illumination intensity along these bodes varies slowly and smoothly, except for the cases of illumination characteristic points used for the first method, and outlines of smooth bodies. This fact results in the split-hair accuracy of the described here first pair of methods with respect to the second pair of methods described below. In case of a small number of the smooth bodies, the second method of tangents can be used only for a validation and an improvement of the method of characteristic points or for only finding a \emph{set} of possible epipolar geometries.

Two more methods are also offered, named CCPM and CTPM, for searching epipolar geometry of images of smooth bodies. The CCPM method is based on search of correspondent points on isophoto curves with the help of correlation of curvatures between these lines. The CTPM method is based on property of the tangential to isophoto curve epipolar line to map into the tangential to correspondent isophoto curves epipolar line. The standard method termed SM and based on knowledge of pairs of the almost exact correspondent points, has been used for testing of these two methods. From discussion of results of these methods in Section ~\ref{sec:implementation} we can conclude that these methods for searching epipolar geometries allow us to find in most cases only a set of the solutions containing the solution close to the true solution, but do not allow us to find this "good" solution among this set. An exception is given for a case when epipoles lie in infinity (that truly, for example, for translation of the camera parallel to plane of the image and in for some other cases). Since epipoles for all solution of chosen set lie approximately on one straight line (a direction of a parallax) in a case epipoles on infinity, their position is determined uniquely by direction of this straight line. The reason of ambiguity in other cases is generally defined by the errors imported by inexactness of assumption of constant brightness. An additional error is also imported with inexactness of a CCPM method, but it is not the main reason of our problem. Among the found (by CCPM method) set almost always there is a solution close to precise, but it can not be found with the help of a CTPM evaluation because of the errors imported by inexactness of assumption of constant brightness.

\section*{Appendix: Setting the partition parameters}

Further we describe how to set the partition parameters, $r_i$ and $\delta\theta_i$, in order to maintain the system resolution and the equal hit properties. We distinguish between three types of regions: regions on the outermost ring, regions within the image and close to the image such that the system resolution condition cannot be maintained and intermediate regions. The parameters depend on the desired system resolution, $\gamma$.

\vspace{-0.4cm}
\subsubsection*{Outer ring regions:}

For regions in the outer ring the epipolar lines are almost parallel. In this case the regions are defined such that, $r_1=\infty$ and $G_{1i}=(\theta_i,\delta\theta,\infty,r_2)$. We next show that it is possible to choose $r_2$ and $\delta\theta$ and $\delta\theta$ such that the region, $G$, satisfies the system resolution.

\begin{figure}[htb]
\centerline{
\psfig{figure=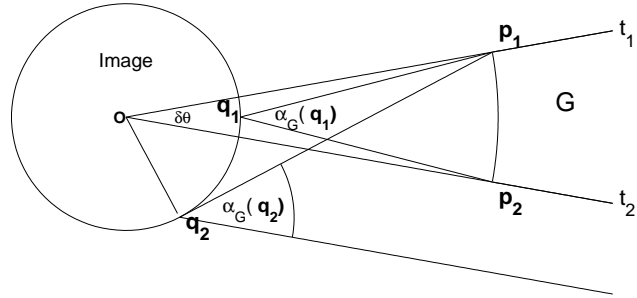,width=3.3in}
}
\caption
{This figure shows a region in the outermost ring, $G$. The maximal angle between an image point and two points in the region, $\alpha_G$, is obtained at one of the two image points ${\bf q_1}$ or ${\bf q_2}$. The point ${\bf q_1}$  is the intersection of the image circle with the angle bisector $\delta\theta$.  The point ${\bf q_2}$  is a point on the image circle such that the tangent to the image circle through ${\bf q_2}$ passes through ${\bf p_1}$.}
\label{fig:outer-ring}
\end{figure}

Using trigonometric considerations, it can be shown that for an outer ring region, $G$, the maximal angle between an image point and two points in the region, $\alpha_G$, is obtained at one of the two image points ${\bf q_1}$ or ${\bf q_2}$ as illustrated in Figure ~\ref{fig:outer-ring}  (the point at which the maximum is obtained depends on $r_2$ and $\delta\theta$).  We next define these points. Let t1 and t2 be the rays that define an outer region, G. Let ${\bf p_1}$ and ${\bf p_2}$ be the delineating points of the region (the intersection points of the region rays and the circle $r_2$). The point ${\bf q_1}$  is the intersection of the image circle with the bisector of the angle $\delta\theta$.  The point ${\bf q_2}$ is the intersection point of the image circle with the tangent from the point ${\bf p_1}$  to the circle . It can be shown using trigonometric considerations that $\alpha_G=max(\alpha_G({\bf q_1}),\alpha_G({\bf q_2}))$ where

\begin{equation}
\alpha_{G}({\bf q_1})=2tg^{-1}({r_2\sin({\delta \theta\over 2})\over r_2\cos({\delta \theta\over 2})-1})  ~~~~{\rm and}~~~~
\alpha_{G}({\bf q_2})=\delta \theta+sin^{-1}(1/r_2)
\label{eq:outer}
\end{equation}

It is therefore possible to set one of the two variables, $r_2$ or $\delta\theta$, and compute the other variable such that $\alpha_{G}<\gamma$ where $\gamma$  is the system resolution. In our system we set $\delta\theta_1$, and solve for $r_2$.  The hit measure, $HM(G)$, is then set to the value computed on $G$  (using formula~\ref{eq:VP2}). This value depends on the initial choice of $\delta\theta_1$.

\vspace{-0.4cm}
\subsubsection*{Intermediate regions:}

Two parameters define an intermediate region, $r_{i+1}$ and $\delta\theta$. We next show how to define these parameters based on the system resolution, $\gamma$ , and the hit measure, $HM(G)$. Similarly to the outer ring case, at an intermediate ring region $G$, $\alpha_G=max(\alpha_G({\bf q_1}),\alpha_G({\bf q_2}))$, where ${\bf q_1}$ is defined as in the outer region case, and ${\bf q_2}$ will be defined next.

\begin{figure}[htb]
\centerline{
\psfig{figure=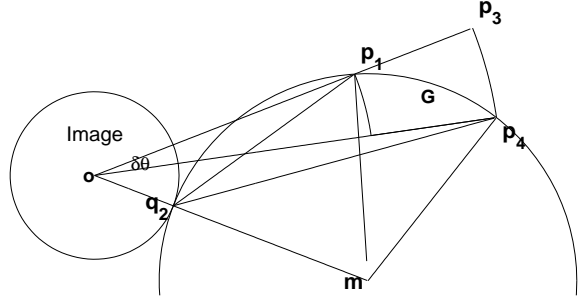,width=3in}
}
\caption
{This figure shows an intermediate region, $G$. The point ${\bf q_2}$  satisfies that the circumscribed circle of the points ${\bf p_1,{\bf p_4}}$ and ${\bf q_2}$ is tangent to the image bounding circle. The angle $\alpha_G({\bf q_2})=  {\bf p_1},{\bf q_1},{\bf p_4}$ is maximal for all image points which are outside the circular section defined by $\delta\theta$.}
\label{fig:interm-ring}
\end{figure}

Let $C_{image}$  be the set of image points on the image bounding circle, outside the image section defined by $\theta_i$ and $\delta\theta$. It can be shown that $\alpha_G({\bf q_2})=\max\{\alpha_G({\bf q})|{\bf q}\in C_{image}\}$,  if the circumscribed circle of the points ${\bf p_1},{\bf p_4}$ and ${\bf q_2}$ is tangent to the image bounding circle (see Figure ~\ref{fig:interm-ring}). This is true since at this point the circumscribed circle has the smaller radius, and therefore the angle $\angle({\bf p_1},{\bf q_1},{\bf p_4})$ is maximal (by the sine theorem). It is now possible to write down the equations that define ${\bf q_2}$.  Let $\vec m$  be the circumscribed circle center of ${\bf q_2},{\bf p_1},{\bf p_4}$.  Assume that the image center is at $(0,0)$ and the image radius is one, in this case, $\norm{{\bf p_1}} = r_{i+1}$, $\norm{{\bf p_4}} = r_i$, and ${\bf q_2}= {{\bf m} \over \norm{{\bf m}}}$.   Under this setup the radius of the circumscribed circle and the angles $\delta\theta$ and $\gamma$ are given by:

\begin{equation}
\begin{array}{rcl}
\cos(\delta\theta)& = & {{\bf p_1} {\bf p_4}^T\over \norm{{\bf p_1}}\norm{{\bf
p_4}}} \cr & & \cr \cos(\gamma) & = & {({\bf p_1}-{\bf q})({\bf p_4}-{\bf
q})^T\over \norm{{\bf p_1}-{\bf q}}\norm{{\bf p_4}-{\bf q}}} \cr & & \cr \norm{{\bf p_1}-{\bf m}}
& = & \norm{{\bf p_4}-{\bf m}} = \norm{{\bf m}}-1\cr
\end{array}
\label{eq:interm}
\end{equation}

Given $r_{i}$ and  $\gamma$ and one of the region parameters $r_{i+1}$ or $\delta\theta$ it is possible to solve these equations for the other parameter, as long as $r_{i+1}$  is large enough. The region must also satisfy the equal hit measure, which is defined in formula~\ref{eq:VP2}.   Our system solves these non linear equations by numerical methods yielding values of the parameters $r_{i+1}$ and $\delta\theta_i$.

\vspace{-0.4cm}
\subsubsection*{Inner regions :}

In the inner regions, the system resolution constraint cannot be maintained. We can only maintain the equal hit measure. We therefore have only one constraint. We set one of the unknown parameters, $r_{i+1}$ and $\delta\theta_i$, arbitrarily, and solve for the other one. In our system we set $\delta\theta_i$ to be a constant equal to the last computed value, and solve for $r_{i+1}$.

\subsection*{Acknowledgment}
We would like to thank Ilan Shimshoni for his supervision and many fruitful ideas used in this paper.

\bibliographystyle{acm}
\bibliography{mybib}
\begin{figure}[htb]
\centerline{
\begin{tabular}{c}
\psfig{figure=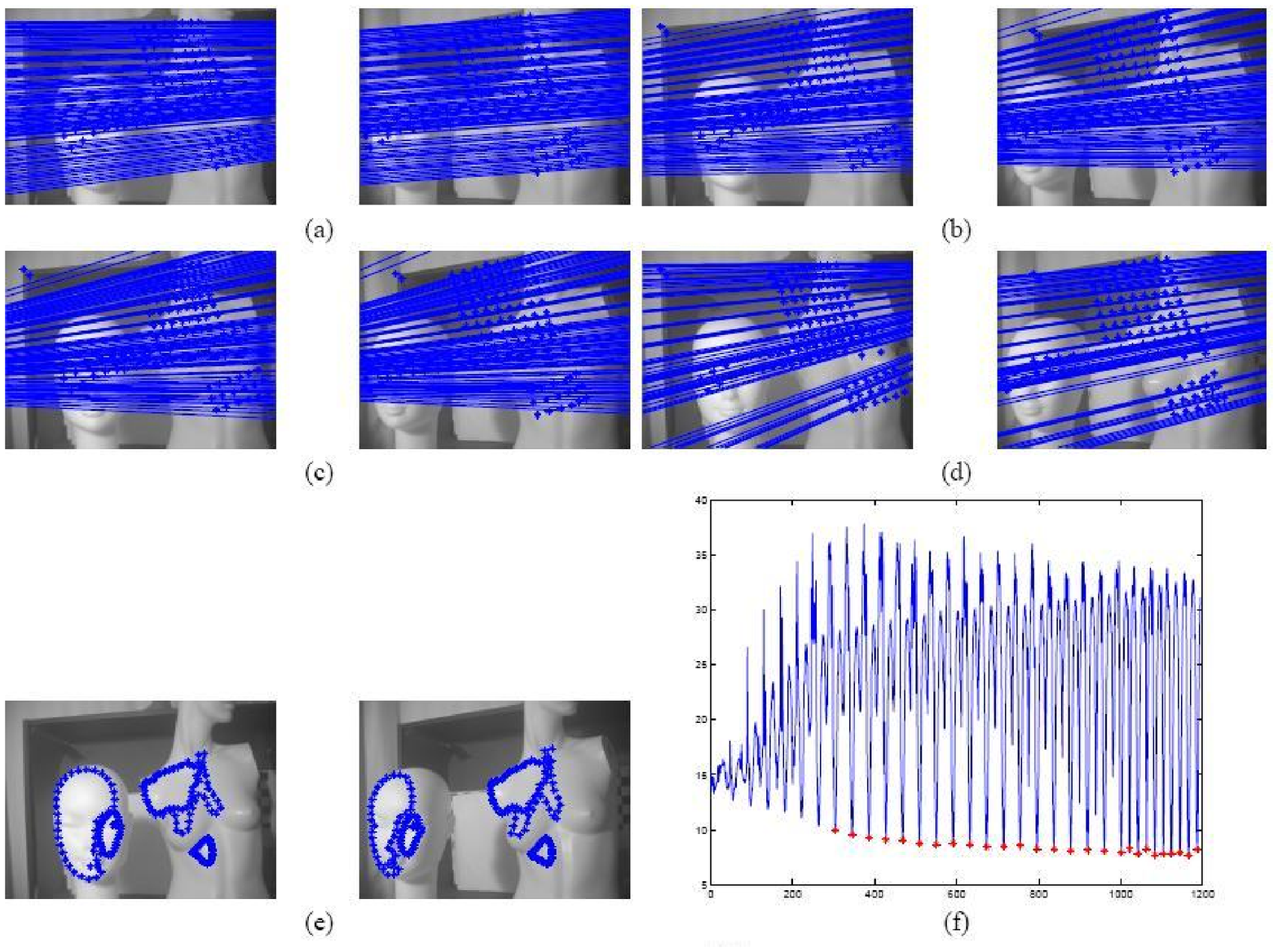,width=5 in}\\
\begin{tabular}{cc}
\psfig{figure=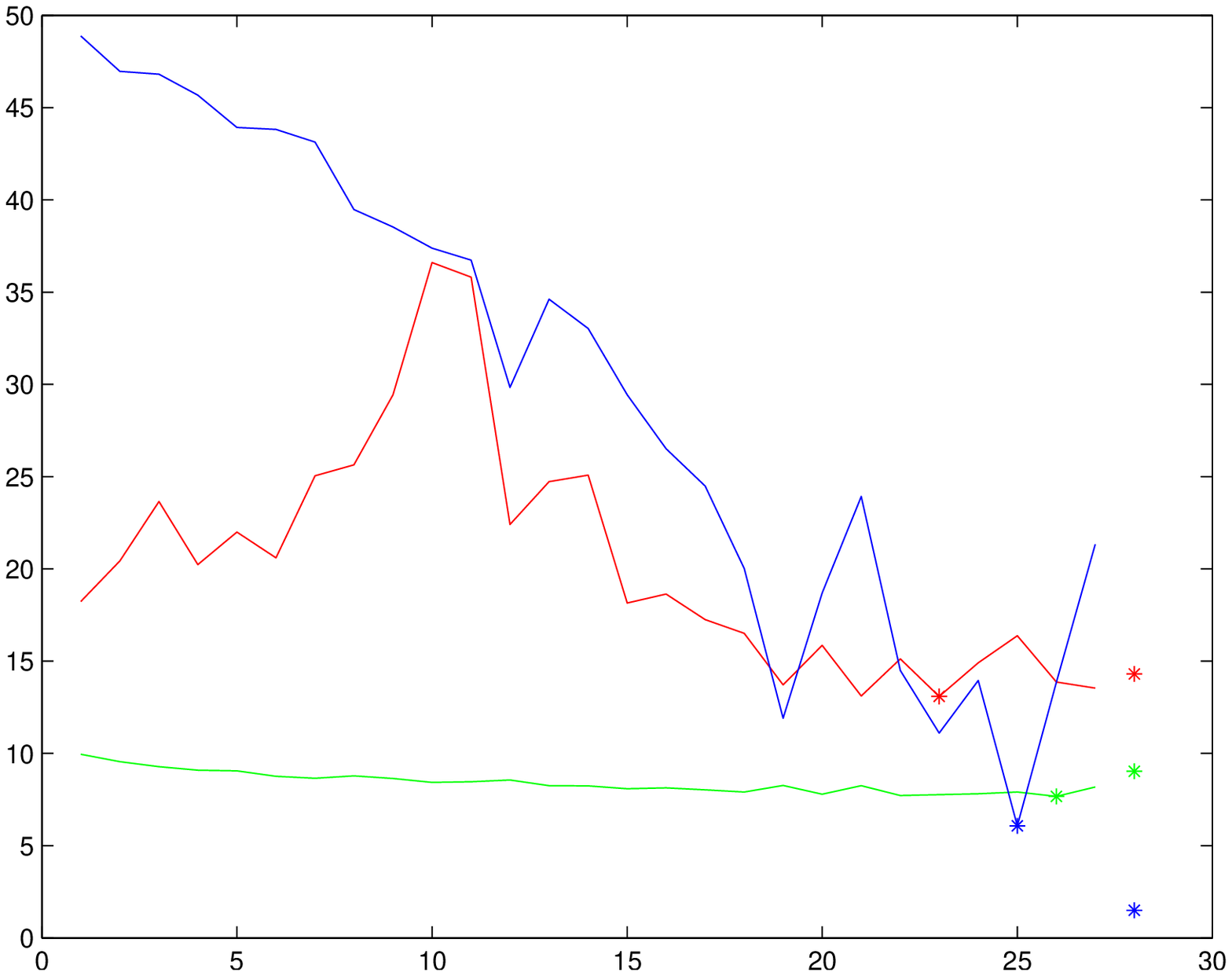,width=2.5 in} &
\psfig{figure=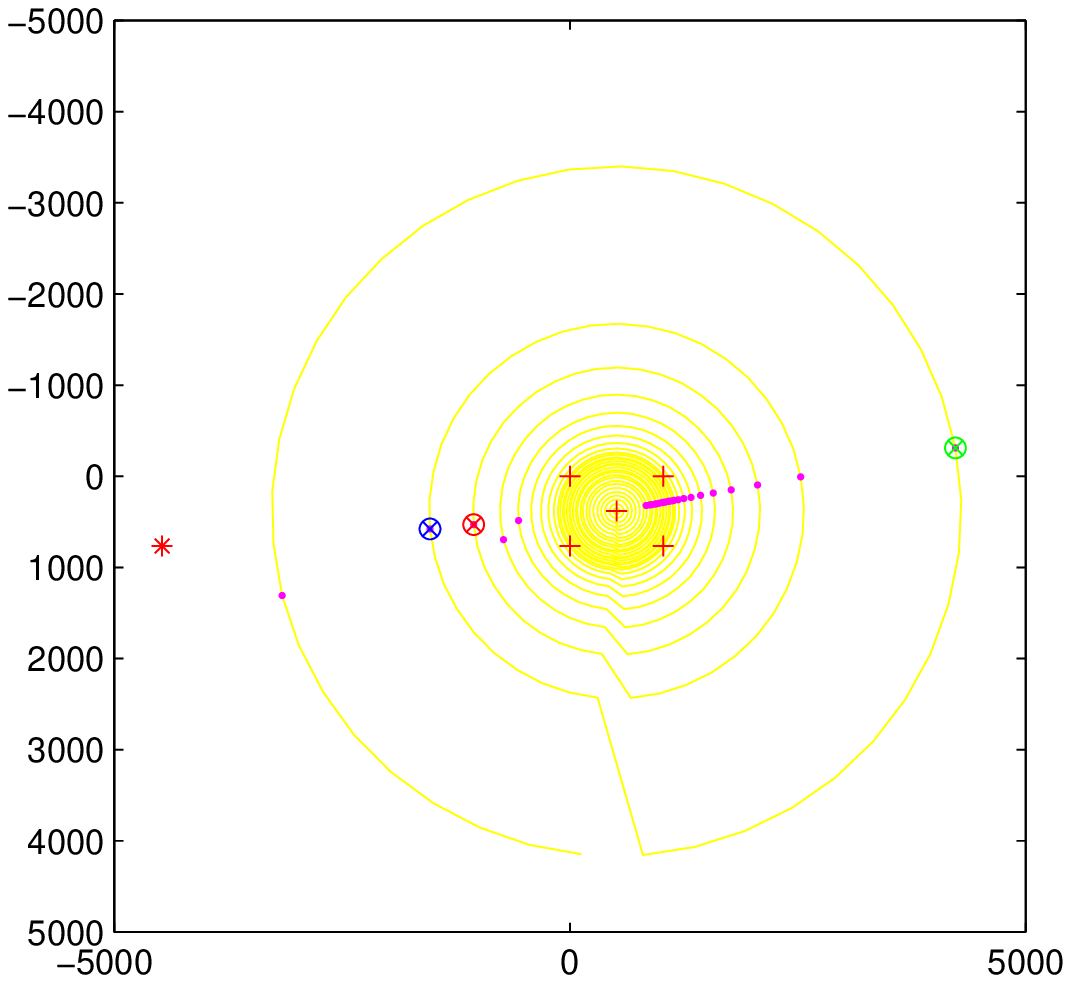,width=2.5 in}\\
(g) & (h)\\
\end{tabular}
\end{tabular}
}
 \caption
{
Case1
 } \label{fig:fige1}
\end{figure}
\begin{figure}[htb]
\centerline{
\begin{tabular}{c}
\psfig{figure=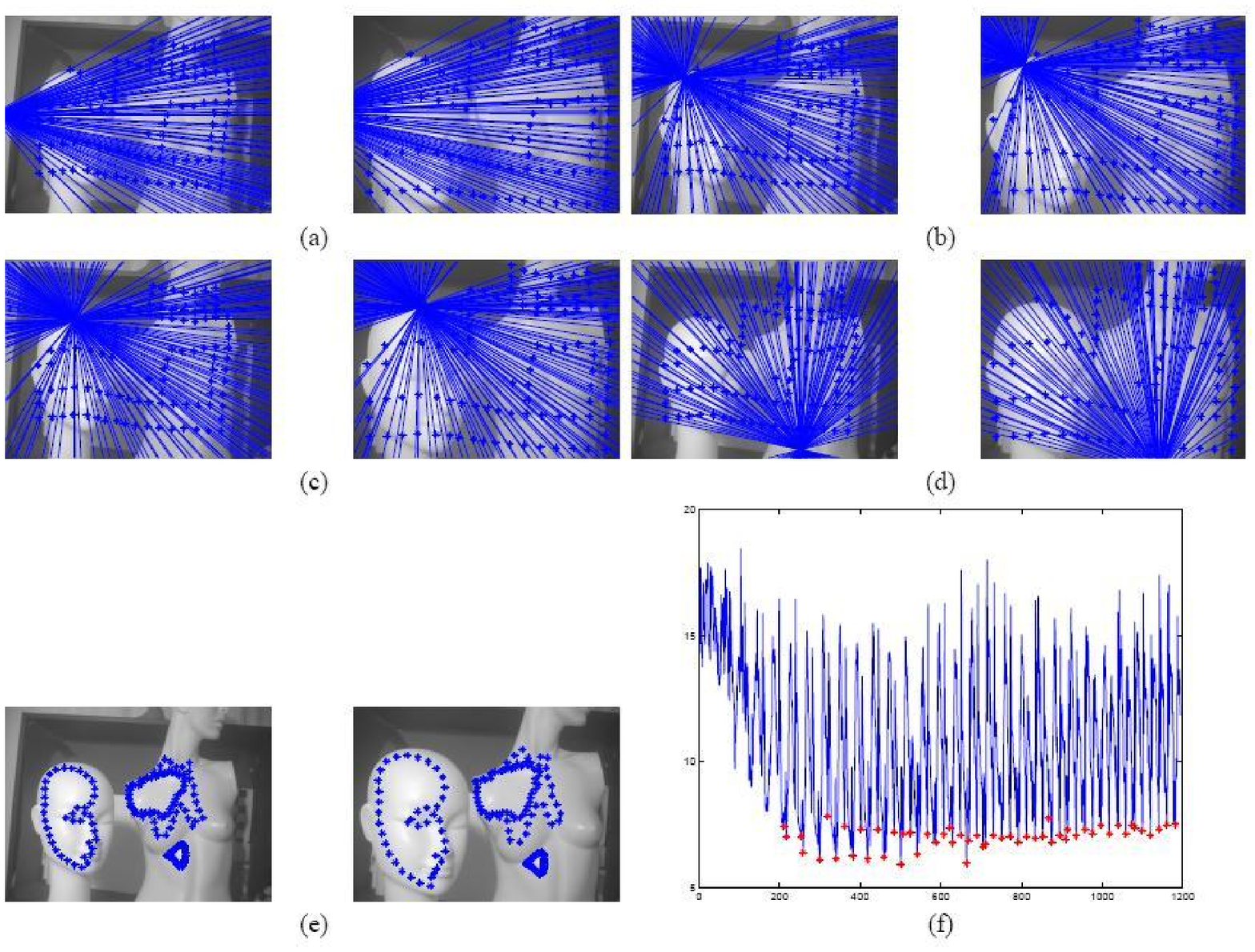,width=5 in}\\
\begin{tabular}{cc}
\psfig{figure=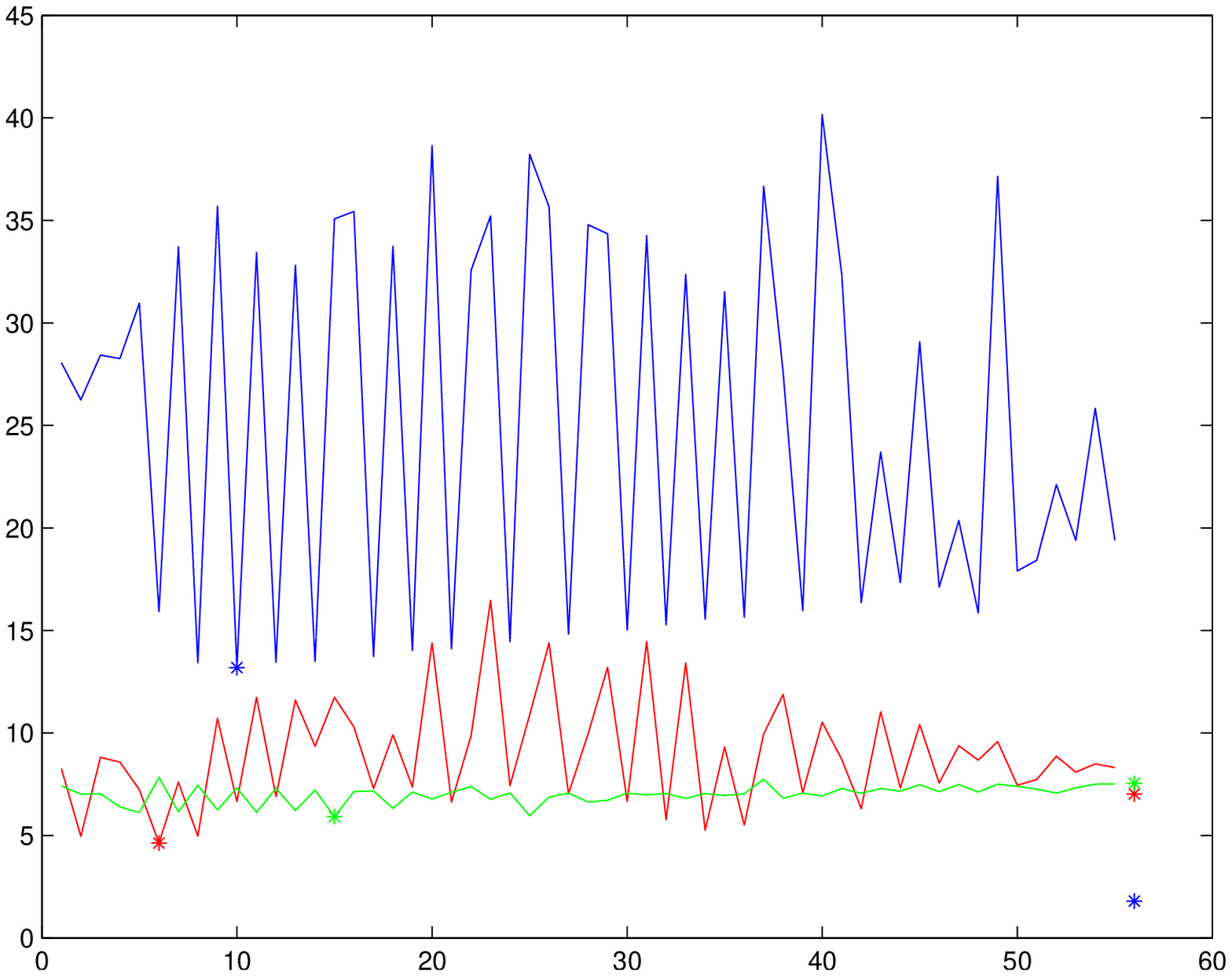,width=2.5 in} &
\psfig{figure=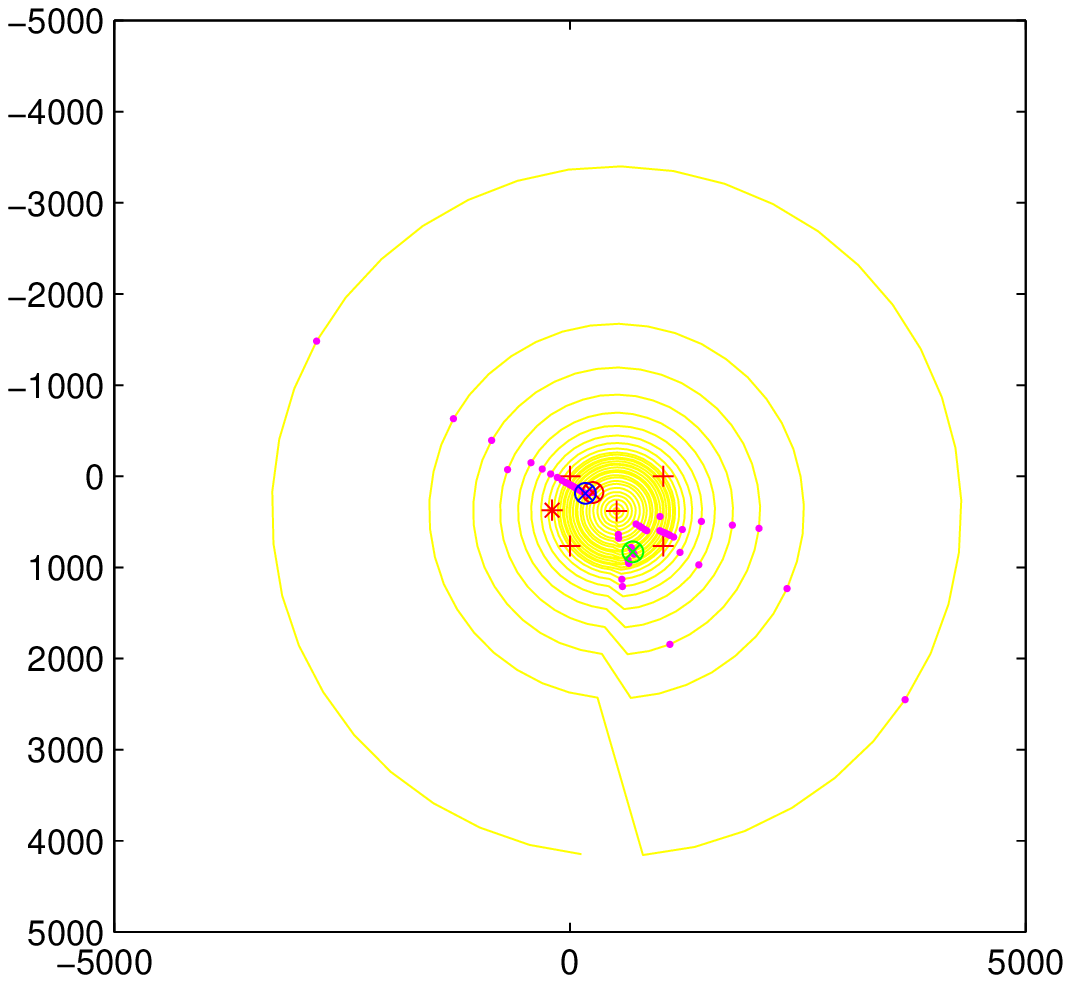,width=2.5 in}\\
(g) & (h)\\
\end{tabular}
\end{tabular}
}
 \caption
{
Case2
 } \label{fig:fige2}
\end{figure}
\begin{figure}[htb]
\centerline{
\begin{tabular}{c}
\psfig{figure=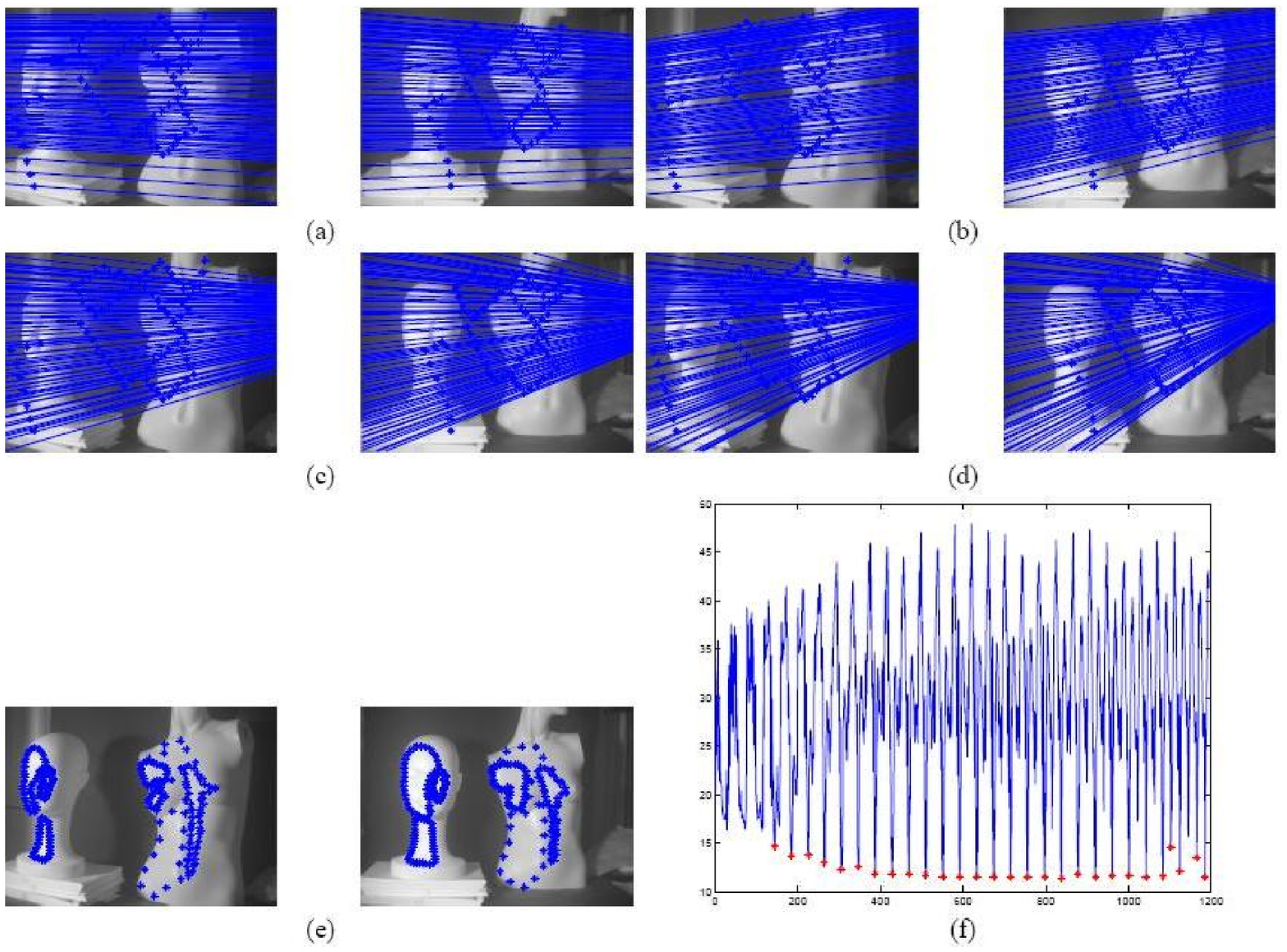,width=5 in}\\
\begin{tabular}{cc}
\psfig{figure=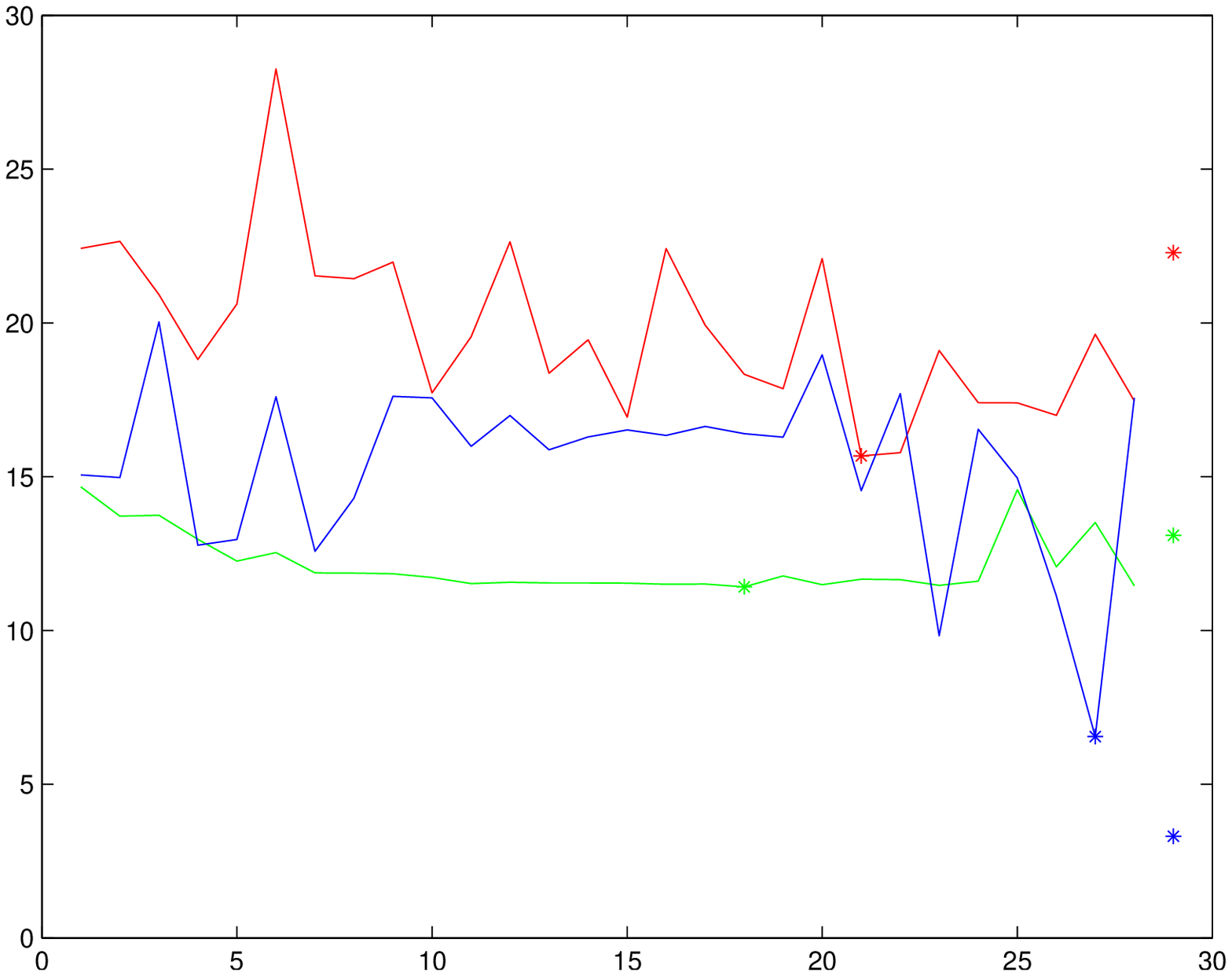,width=2.5 in} &
\psfig{figure=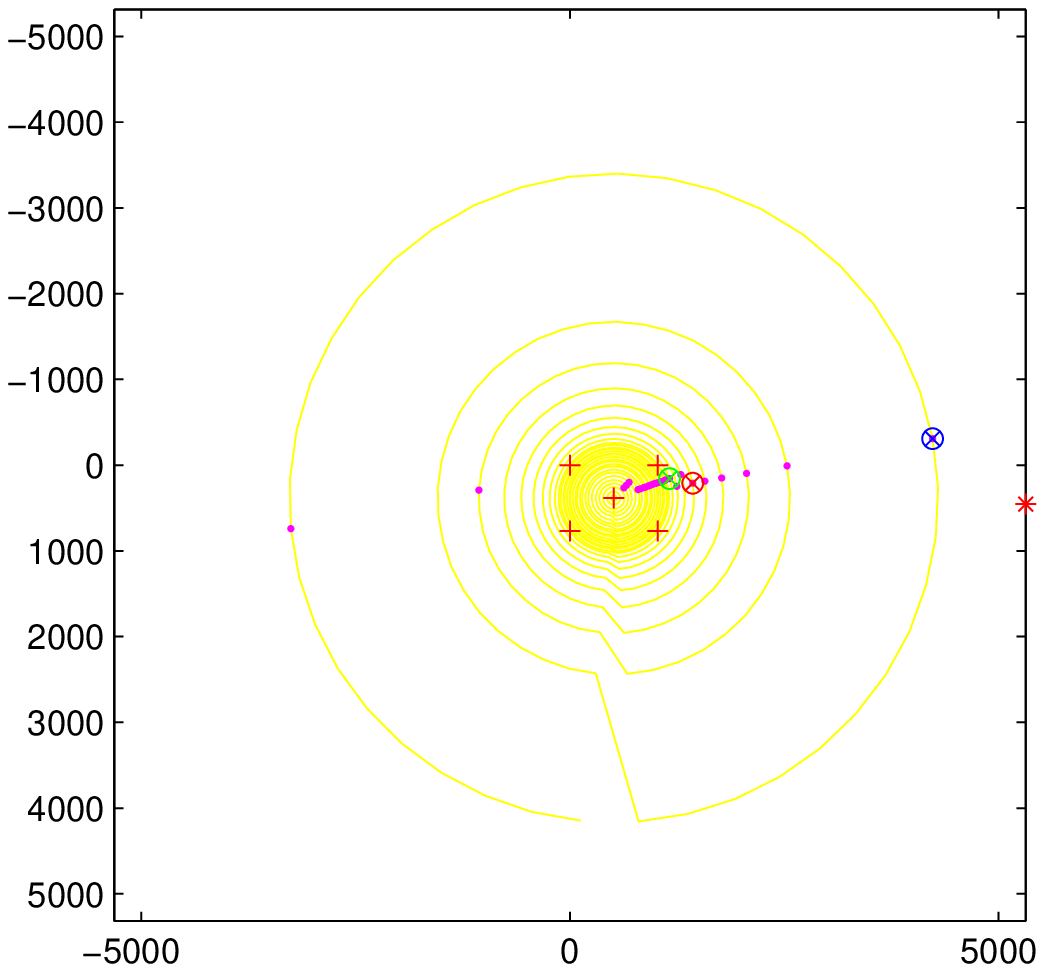,width=2.5 in}\\
(g) & (h)\\
\end{tabular}
\end{tabular}
}
 \caption
{
Case3a
 } \label{fig:fige3a}
\end{figure}
\begin{figure}[htb]
\centerline{
\begin{tabular}{c}
\psfig{figure=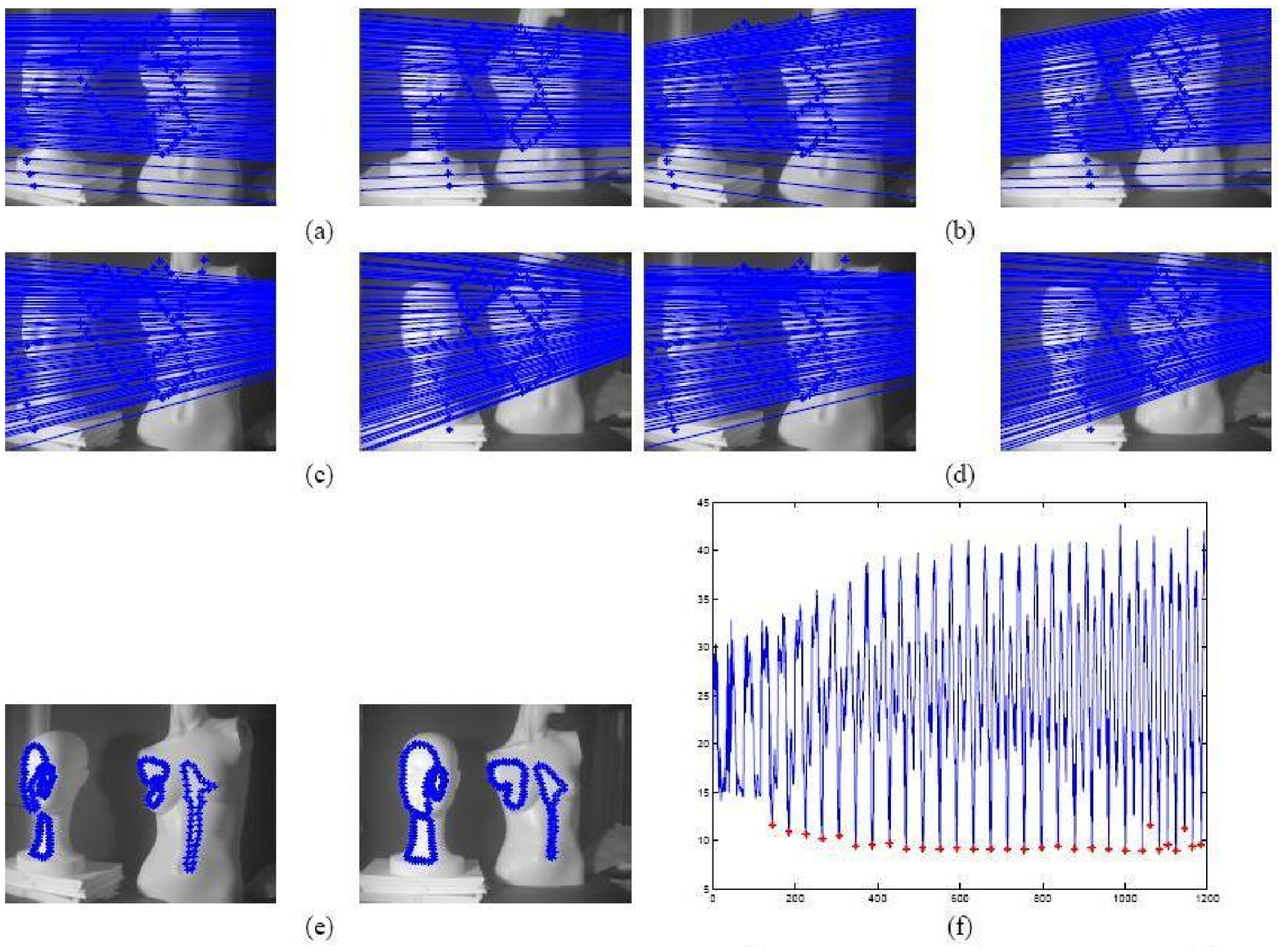,width=5 in}\\
\begin{tabular}{cc}
\psfig{figure=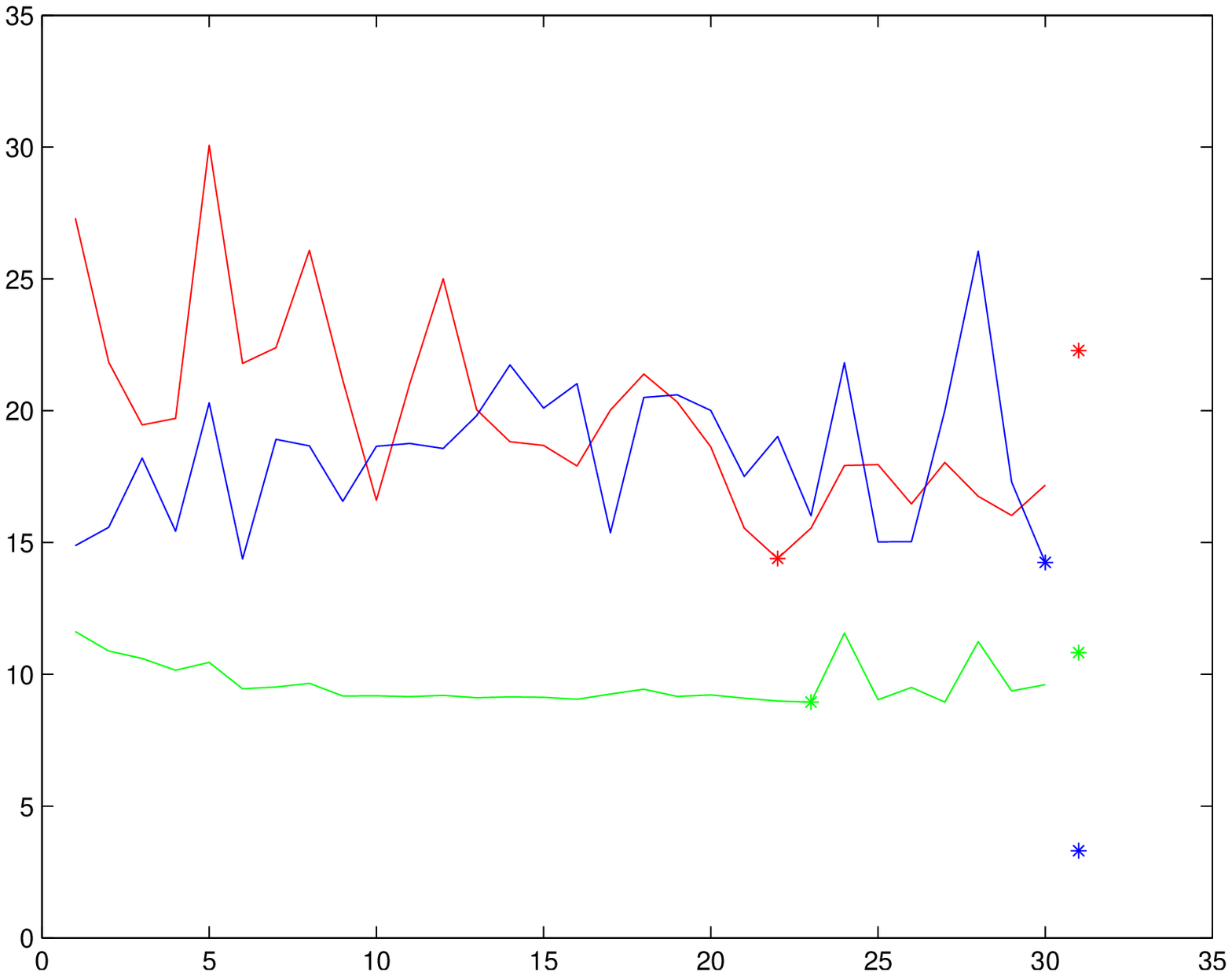,width=2.5 in} &
\psfig{figure=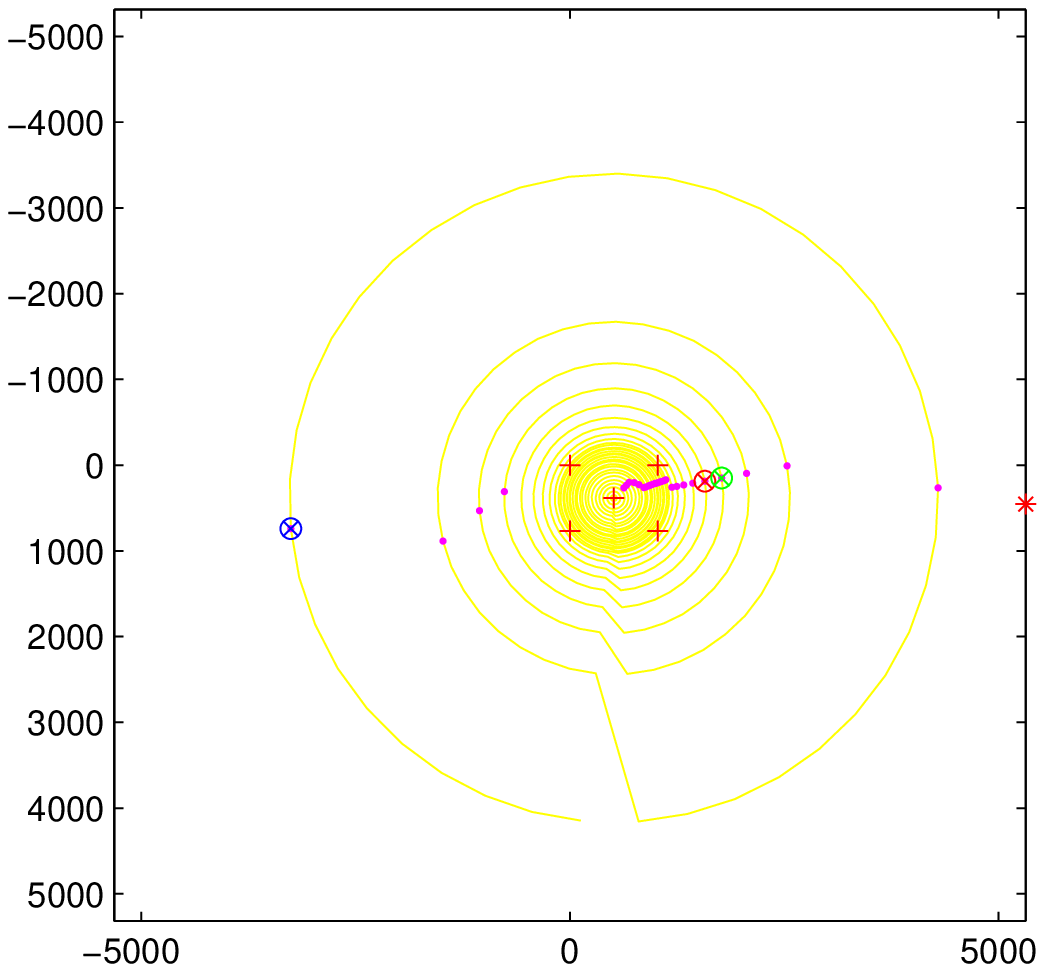,width=2.5 in}\\
(g) & (h)\\
\end{tabular}
\end{tabular}
}
 \caption
{
Case3b
 } \label{fig:fige3b}
\end{figure}
\begin{figure}[htb]
\centerline{
\begin{tabular}{c}
\psfig{figure=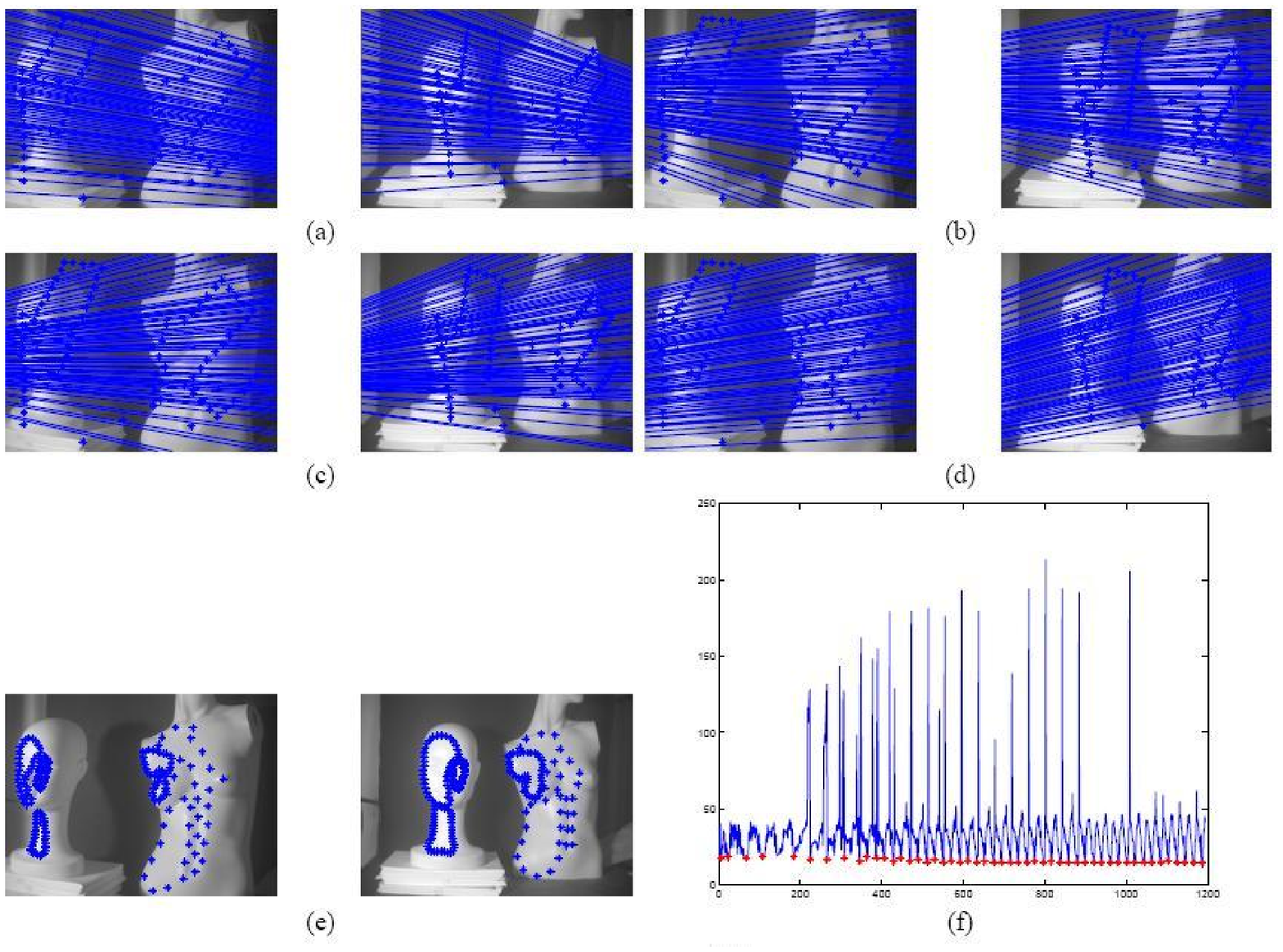,width=5 in}\\
\begin{tabular}{cc}
\psfig{figure=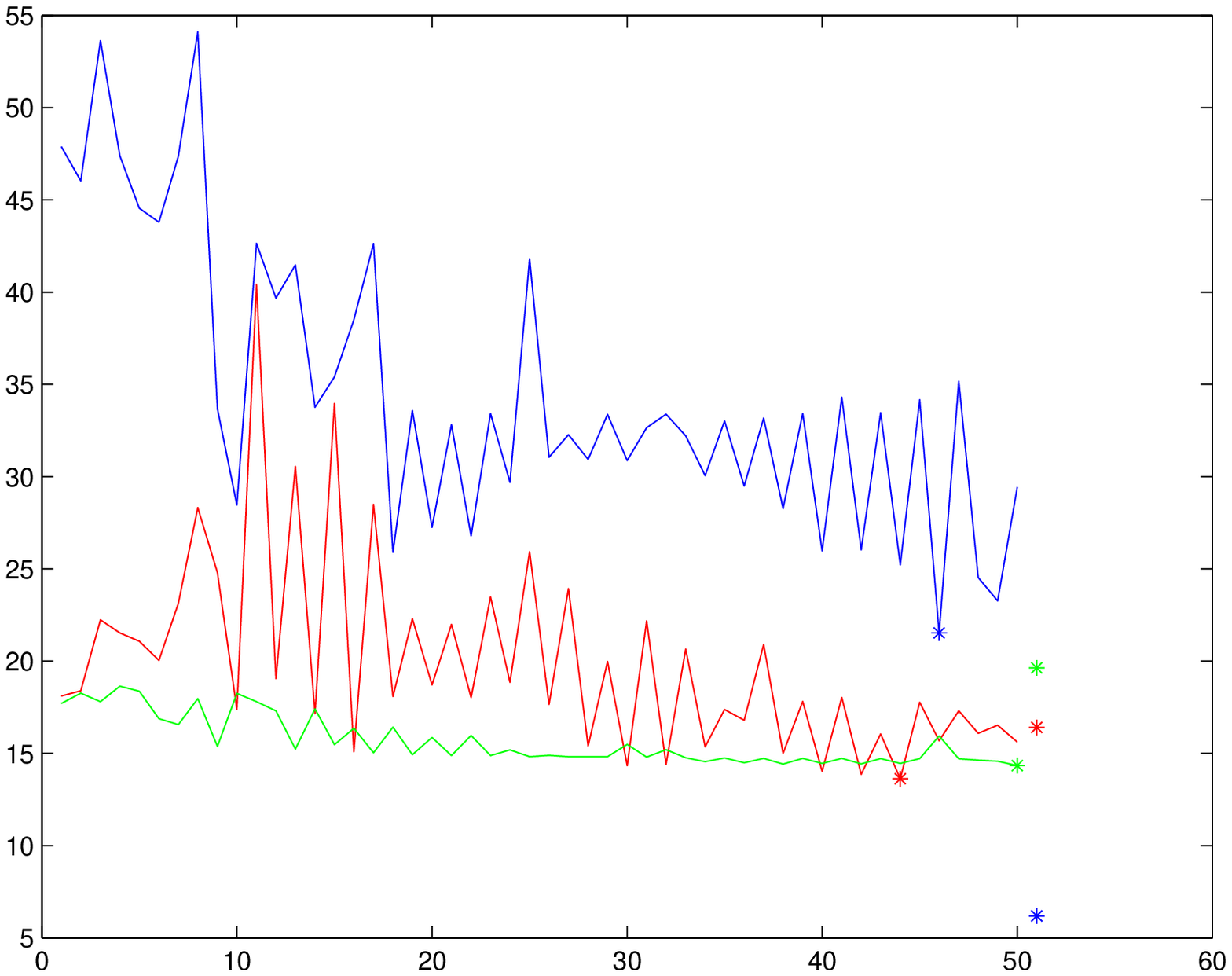,width=2.5 in} &
\psfig{figure=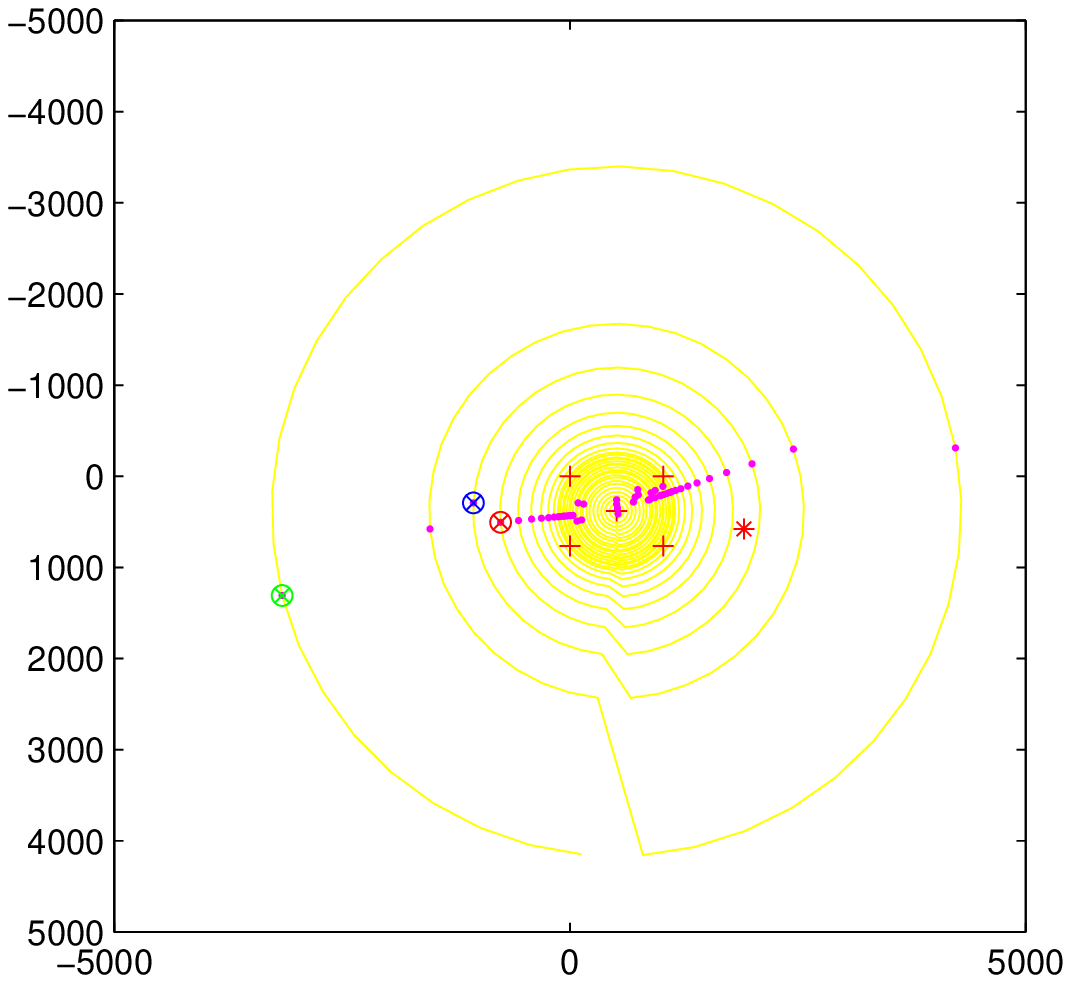,width=2.5 in}\\
(g) & (h)\\
\end{tabular}
\end{tabular}
}
 \caption
{
Case4a
 } \label{fig:fige4a}
\end{figure}
\begin{figure}[htb]
\centerline{
\begin{tabular}{c}
\psfig{figure=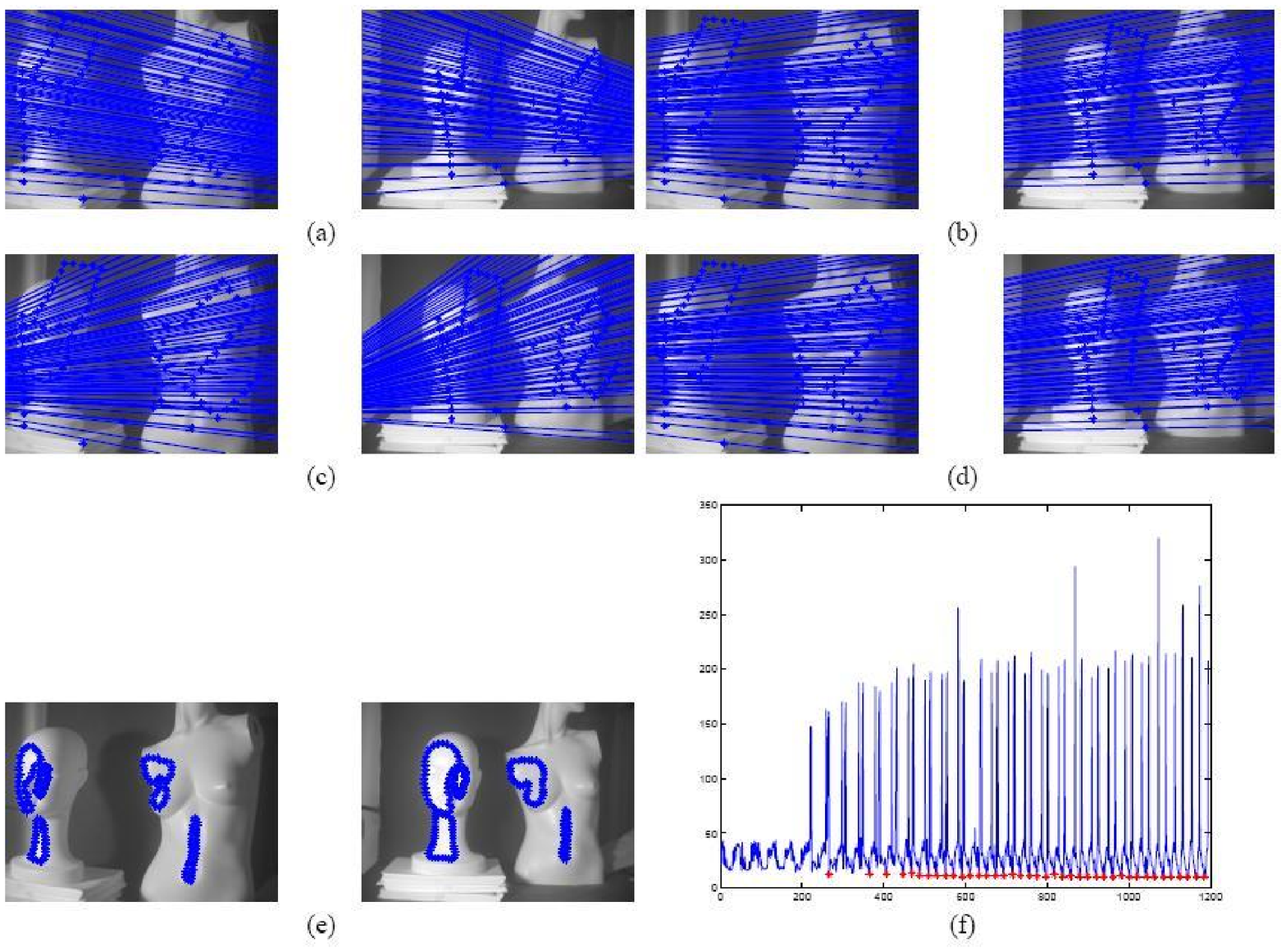,width=5 in}\\
\begin{tabular}{cc}
\psfig{figure=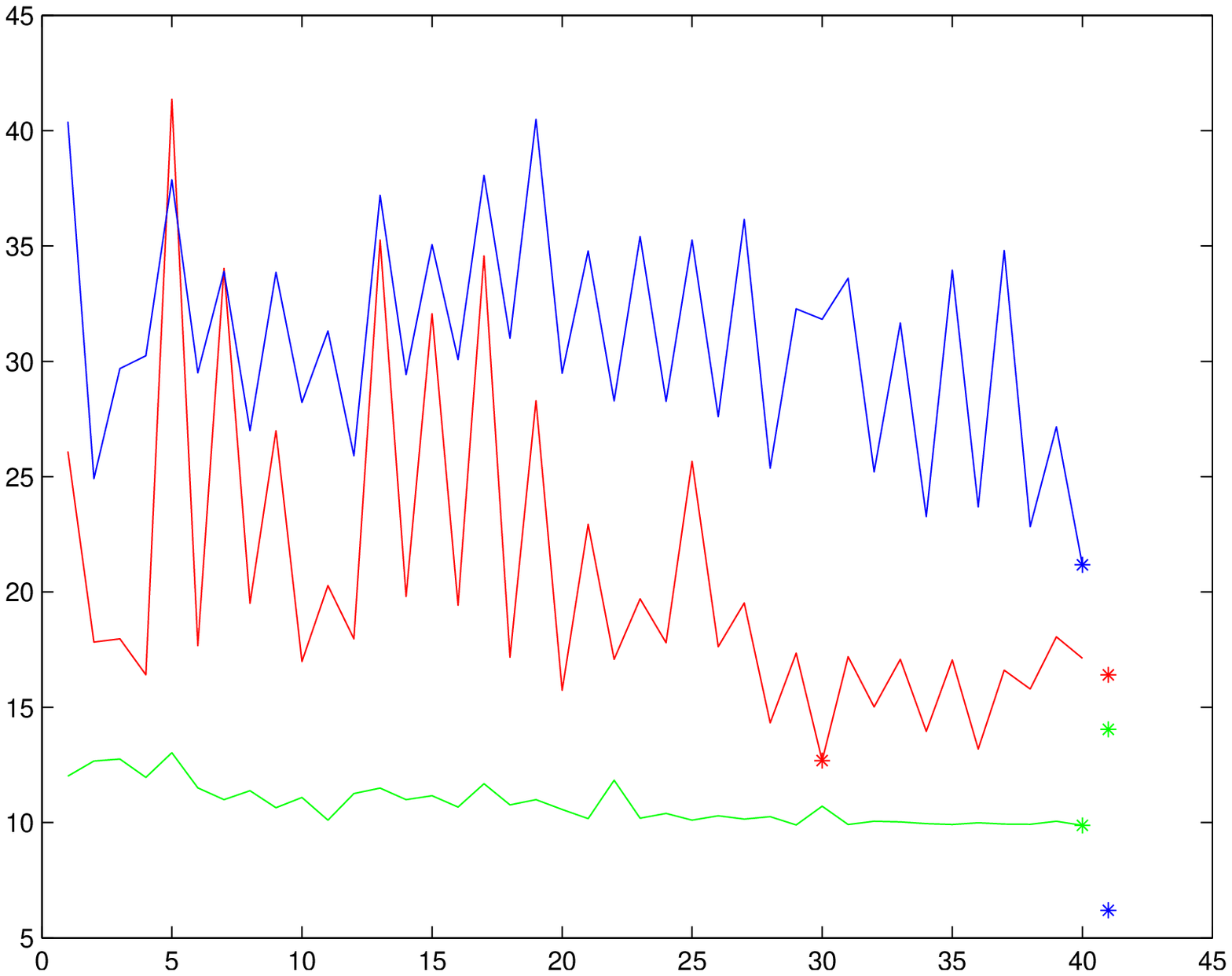,width=2.5 in} &
\psfig{figure=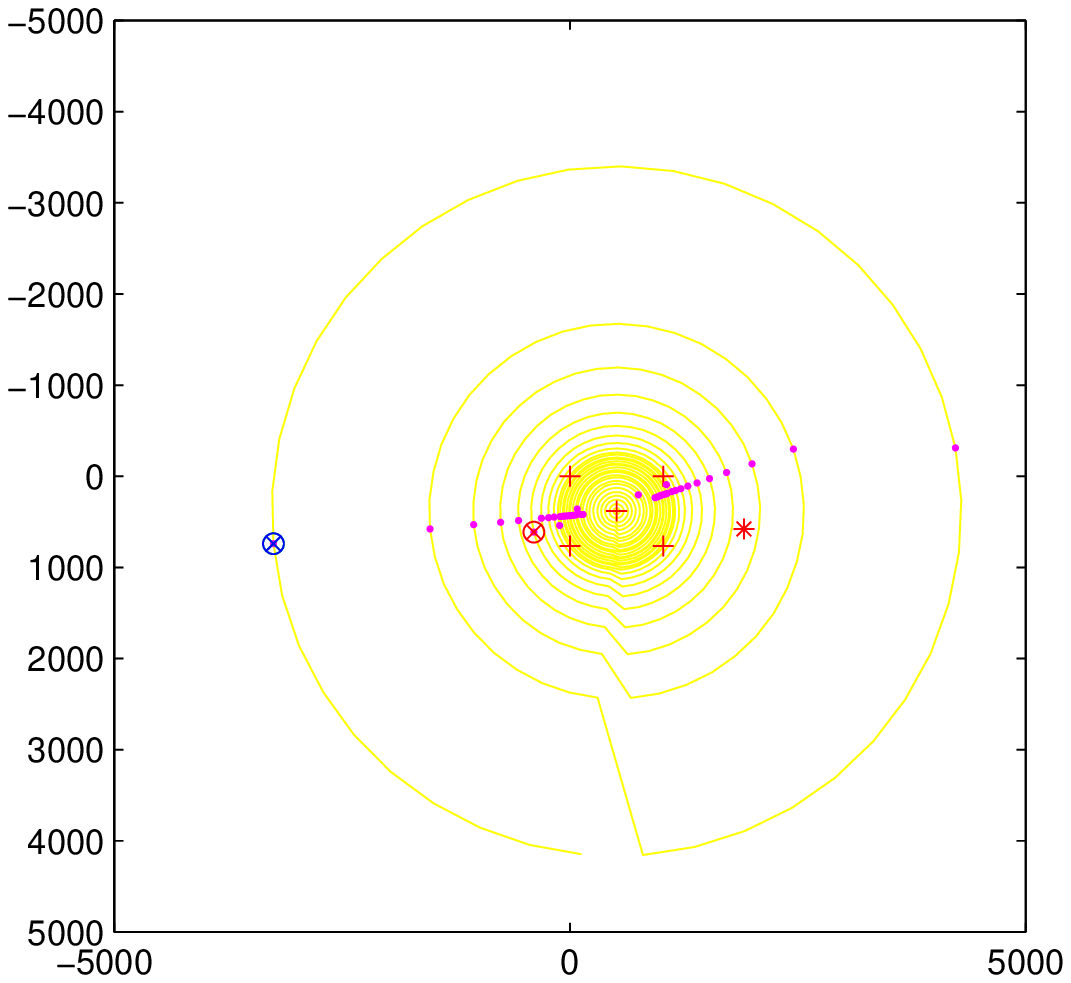,width=2.5 in}\\
(g) & (h)\\
\end{tabular}
\end{tabular}
}
 \caption
{
 Case4b
 } \label{fig:fige4b}
\end{figure}
\begin{figure}[htb]
\centerline{
\begin{tabular}{c}
\psfig{figure=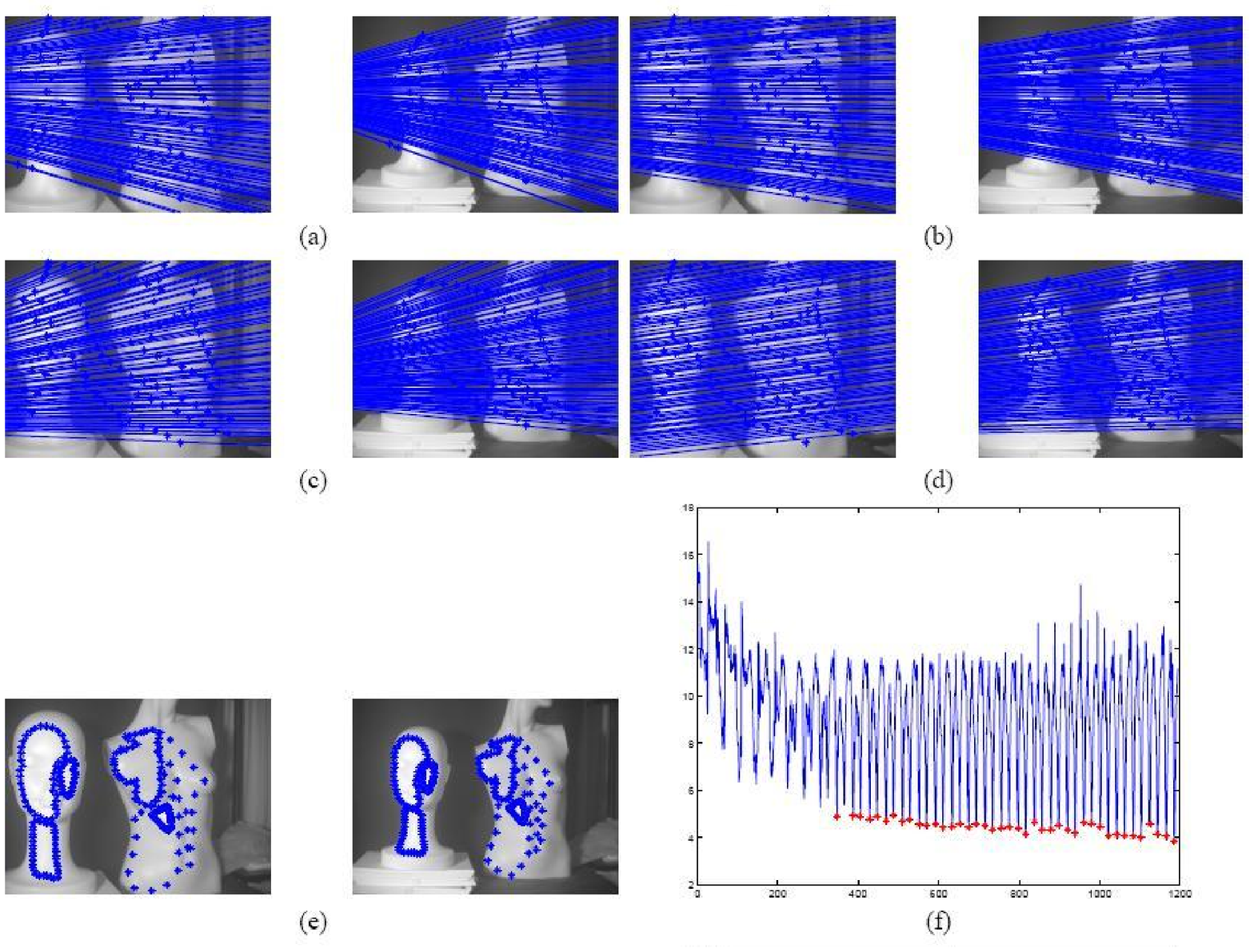,width=5 in}\\
\begin{tabular}{cc}
\psfig{figure=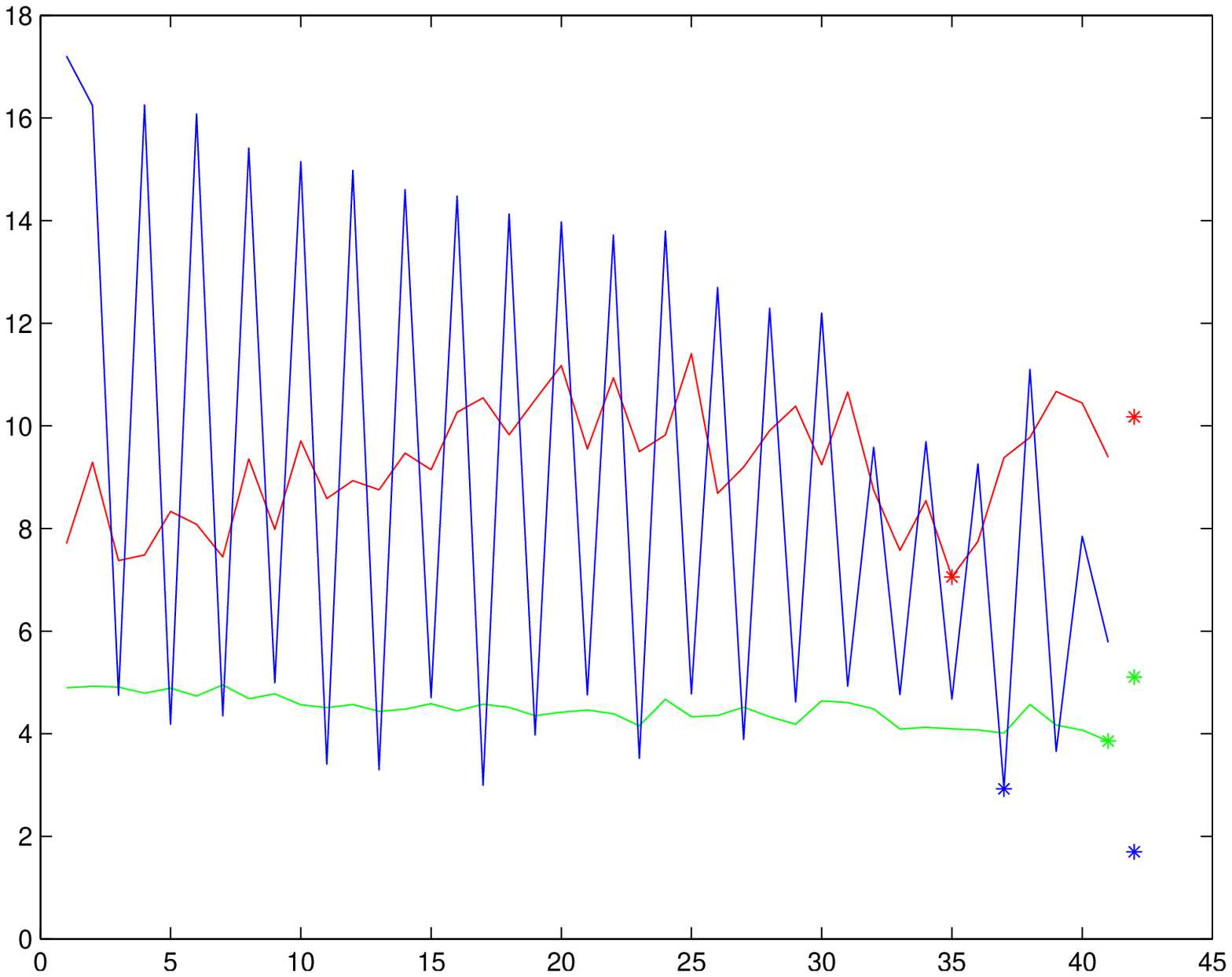,width=2.5 in} &
\psfig{figure=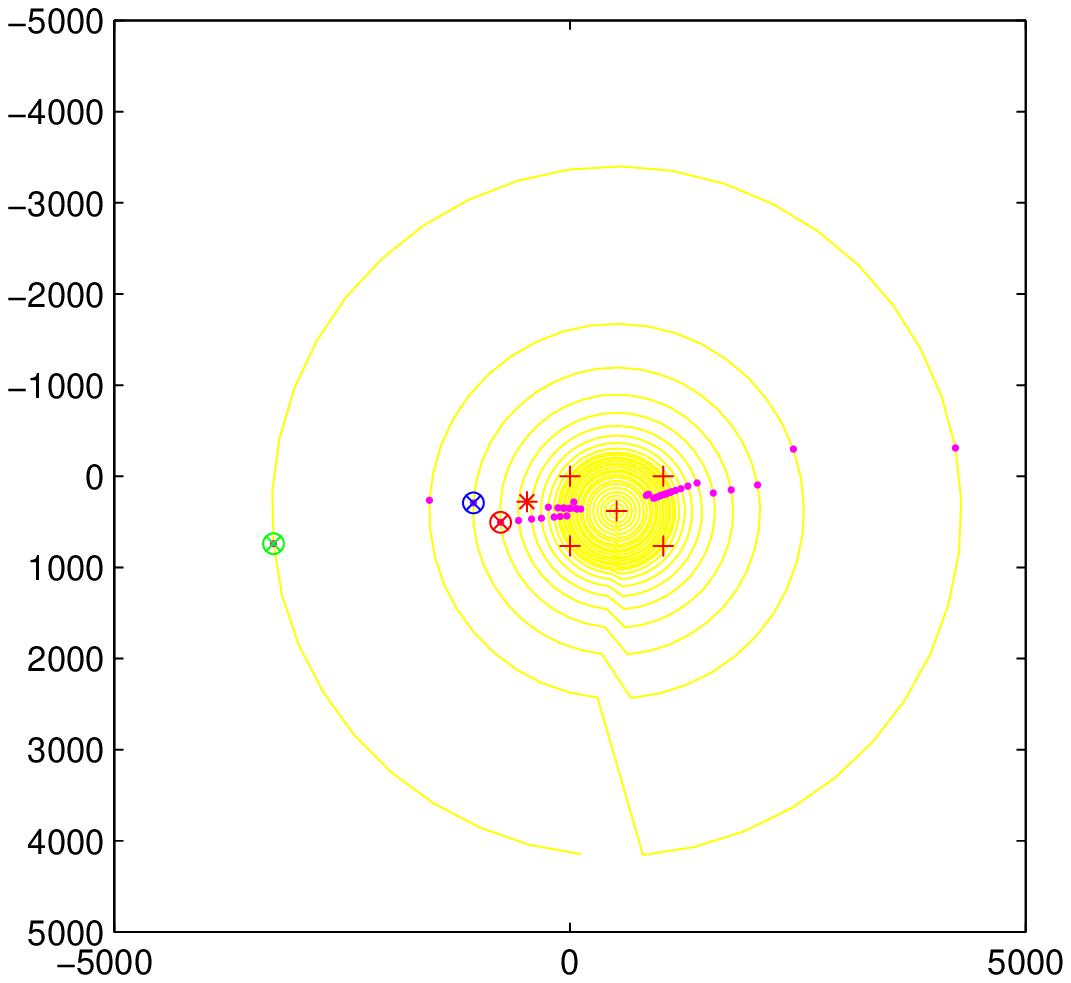,width=2.5 in}\\
(g) & (h)\\
\end{tabular}
\end{tabular}
}
 \caption
{
Case5
 } \label{fig:fige5}
\end{figure}
\begin{figure}[htb]
\centerline{
\begin{tabular}{c}
\psfig{figure=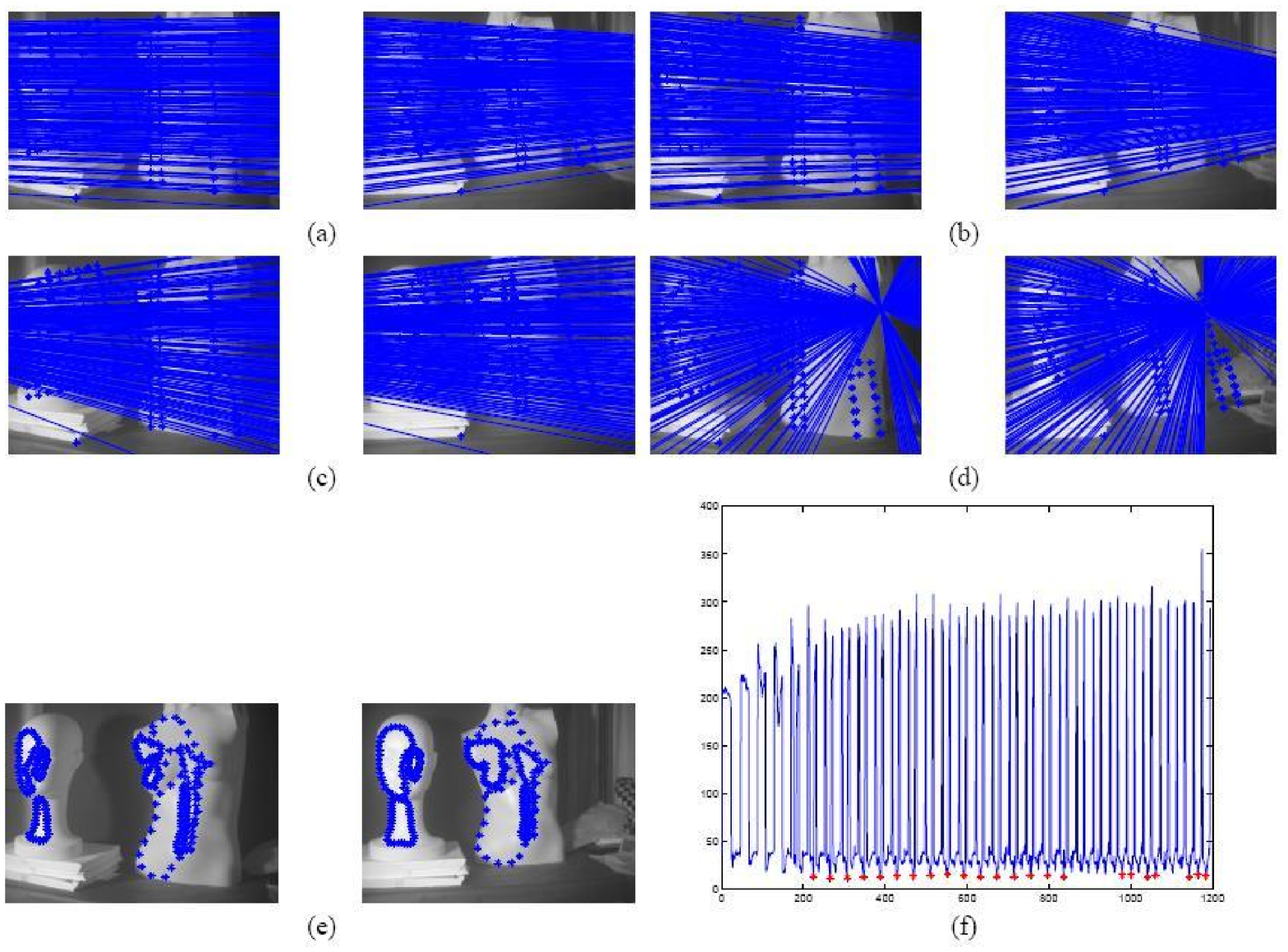,width=5 in}\\
\begin{tabular}{cc}
\psfig{figure=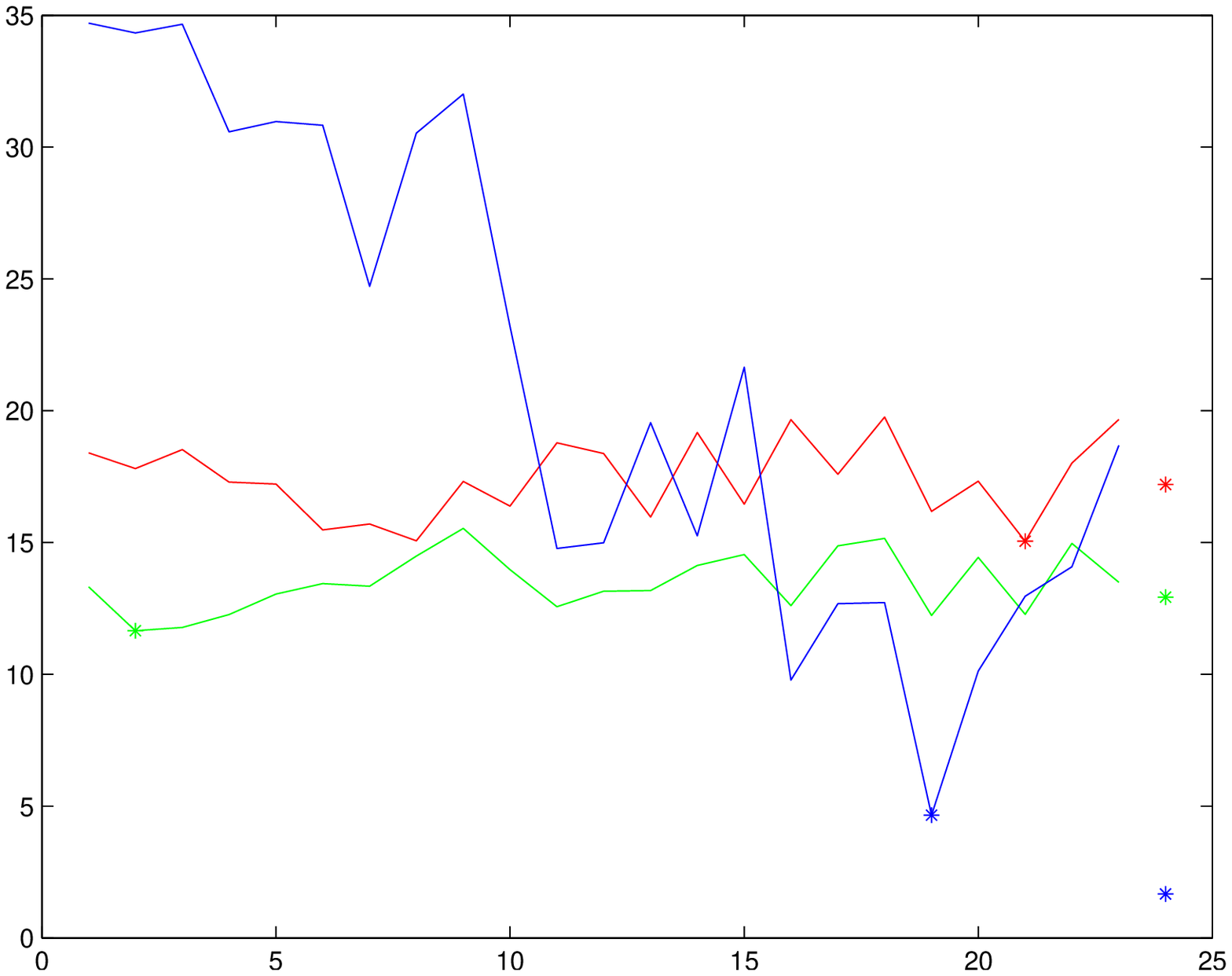,width=2.5 in} &
\psfig{figure=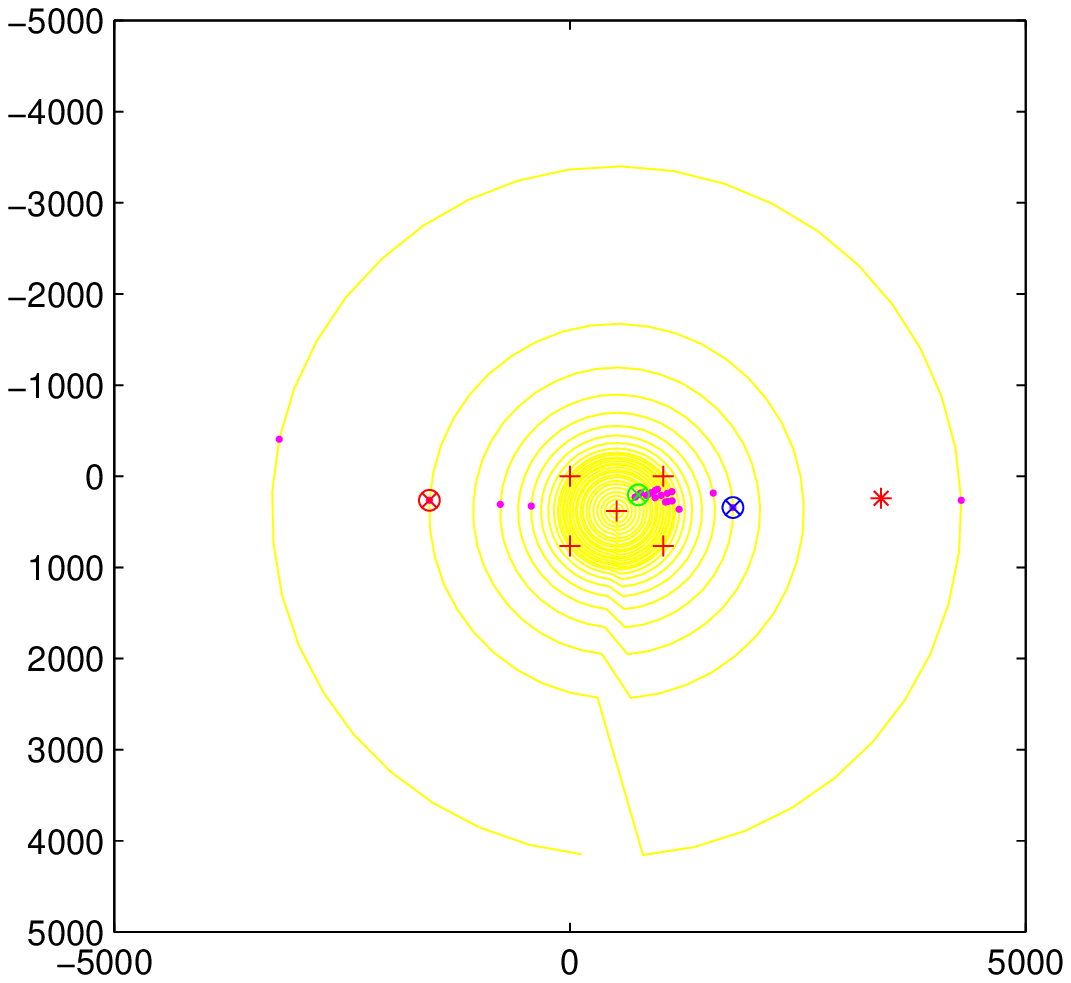,width=2.5 in}\\
(g) & (h)\\
\end{tabular}
\end{tabular}
}
 \caption
{
Case6a
 } \label{fig:fige6a}
\end{figure}
\begin{figure}[htb]
\centerline{
\begin{tabular}{c}
\psfig{figure=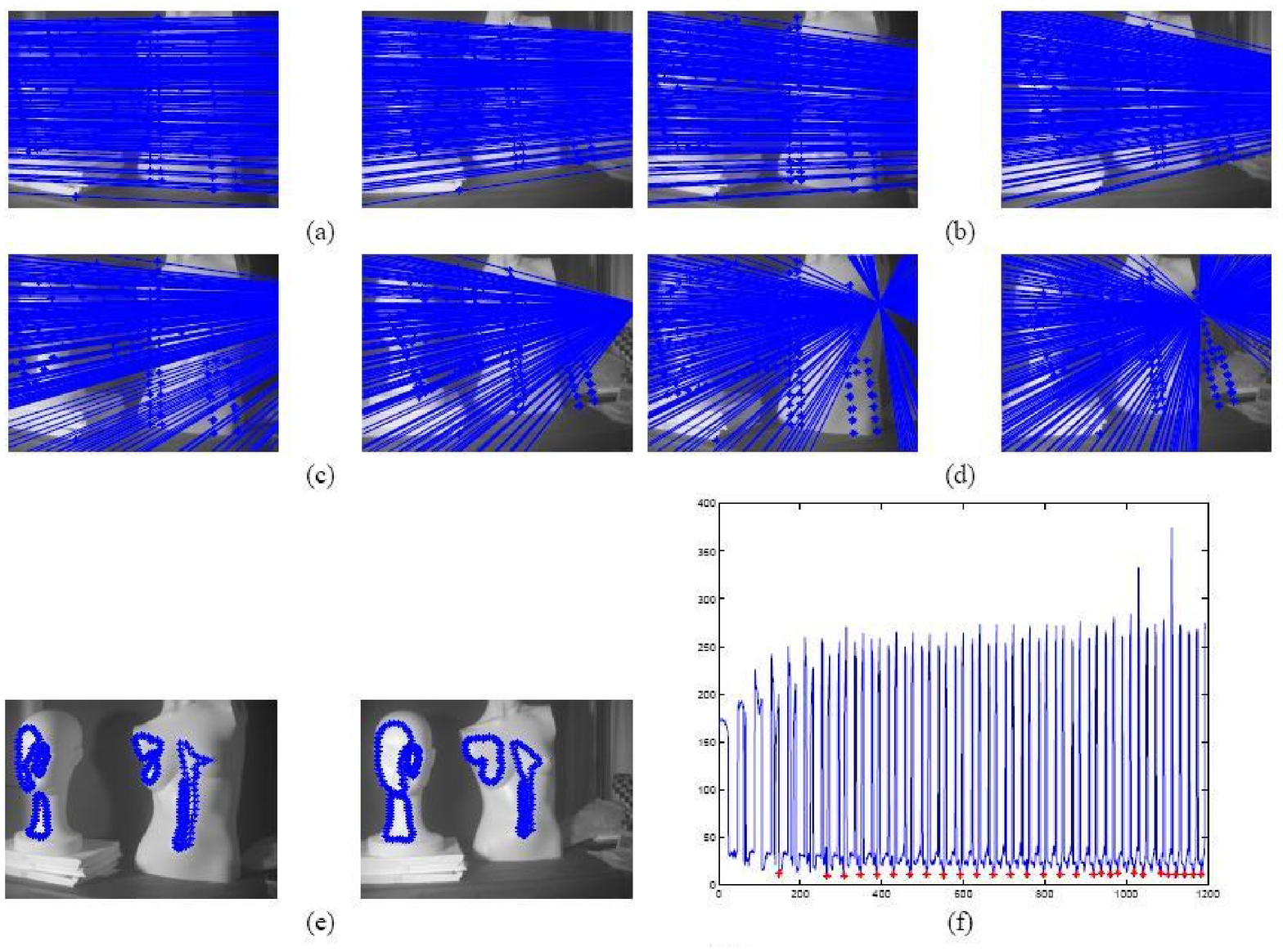,width=5 in}\\
\begin{tabular}{cc}
\psfig{figure=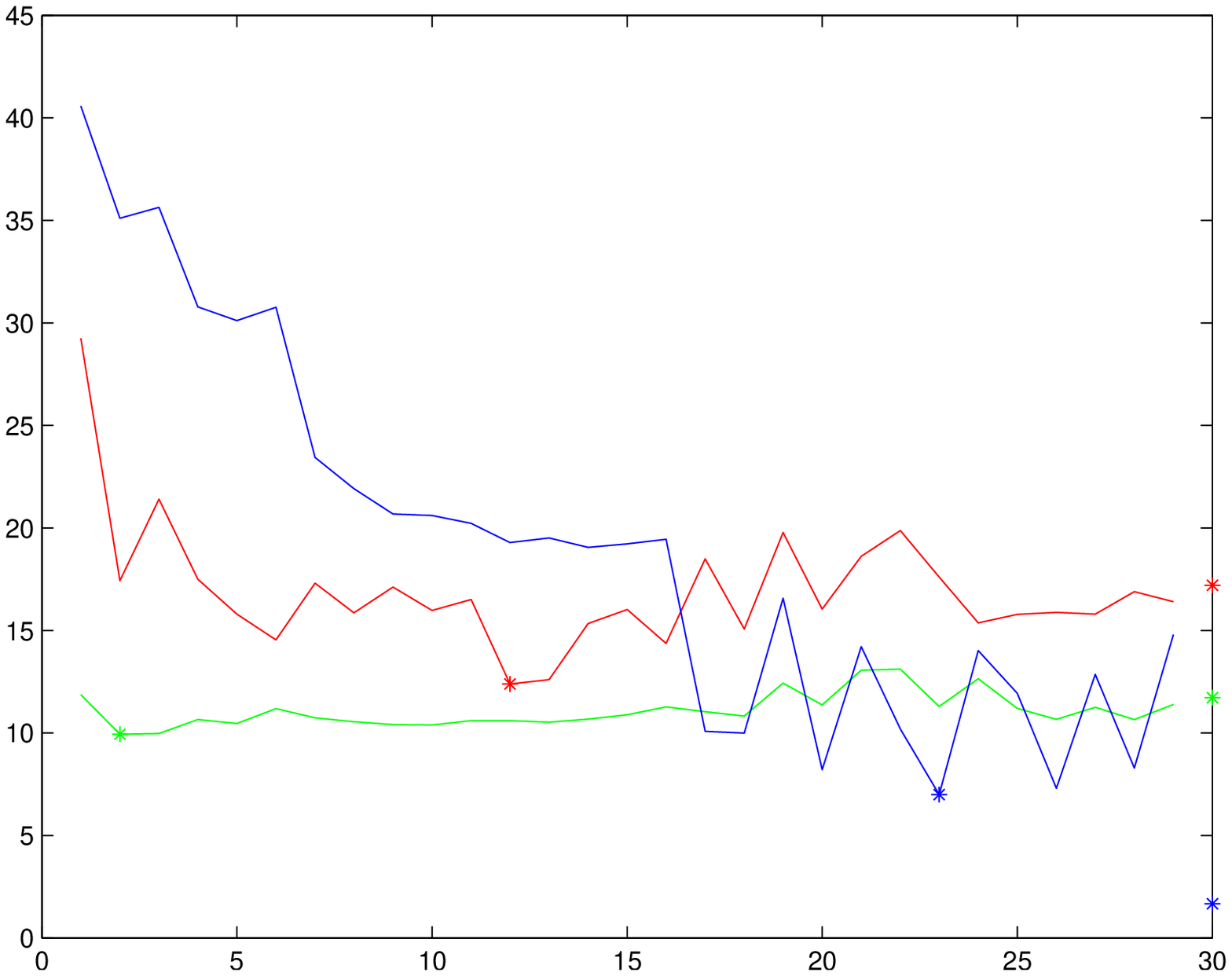,width=2.5 in} &
\psfig{figure=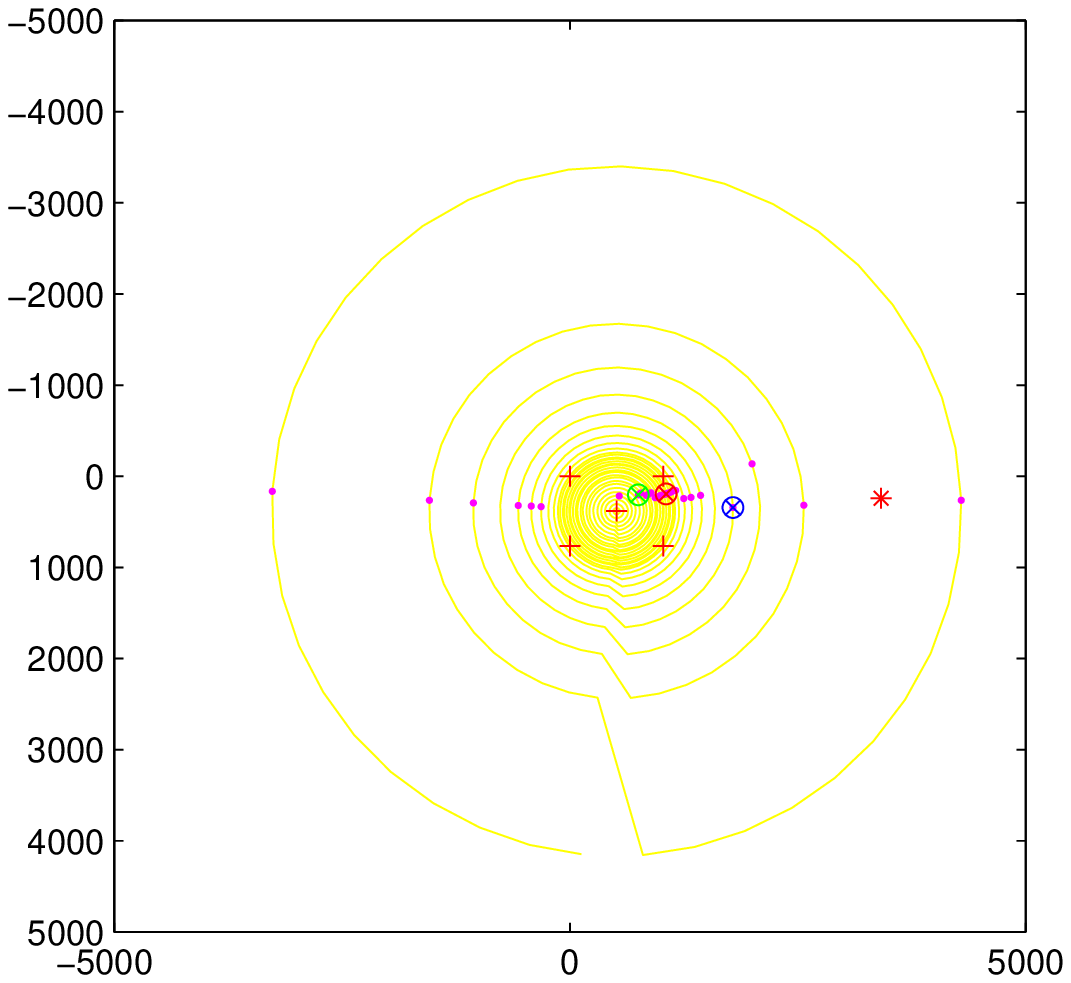,width=2.5 in}\\
(g) & (h)\\
\end{tabular}
\end{tabular}
}
 \caption
{
Case6b
 } \label{fig:fige6b}
\end{figure}
\begin{figure}[htb]
\centerline{
\begin{tabular}{c}
\psfig{figure=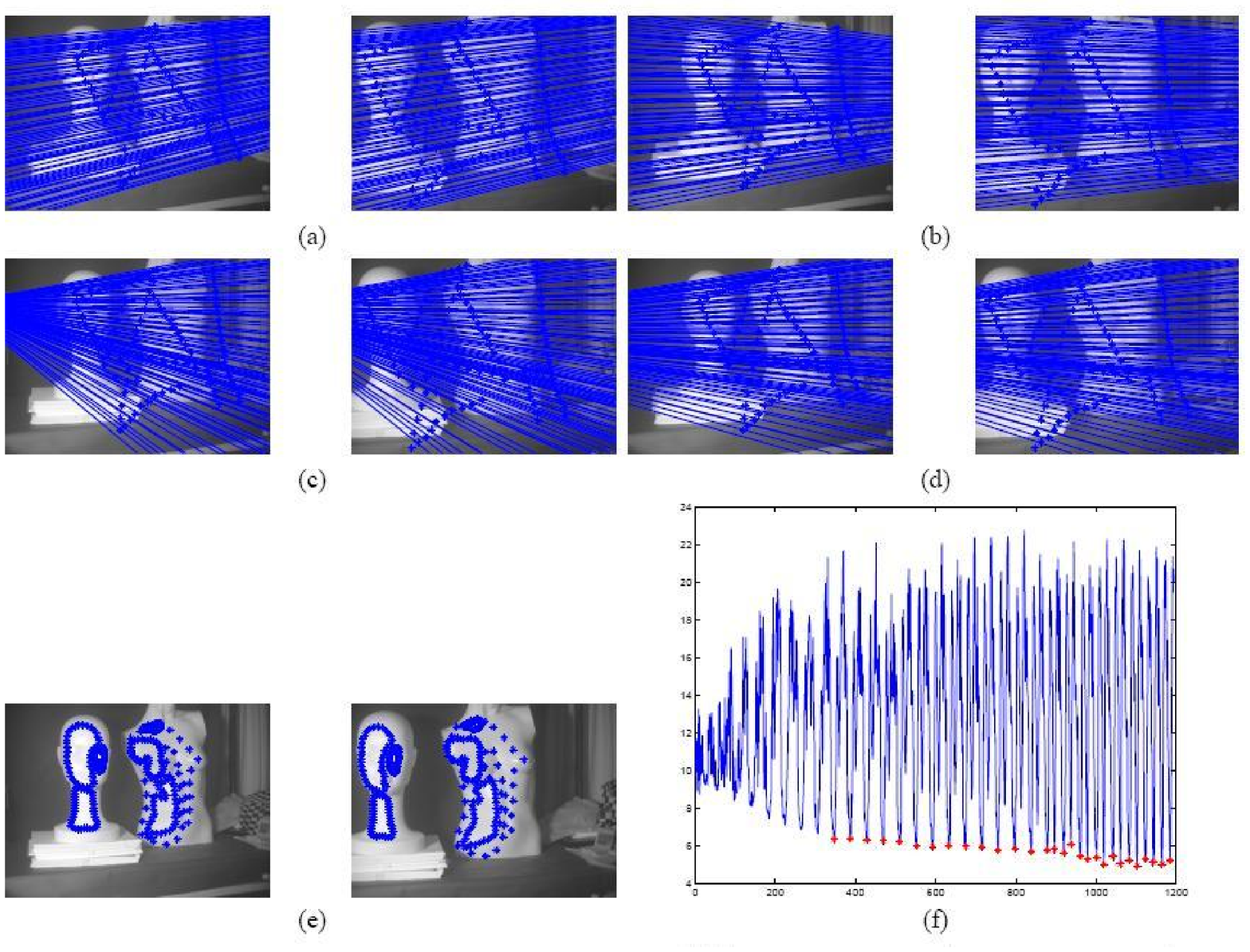,width=5 in}\\
\begin{tabular}{cc}
\psfig{figure=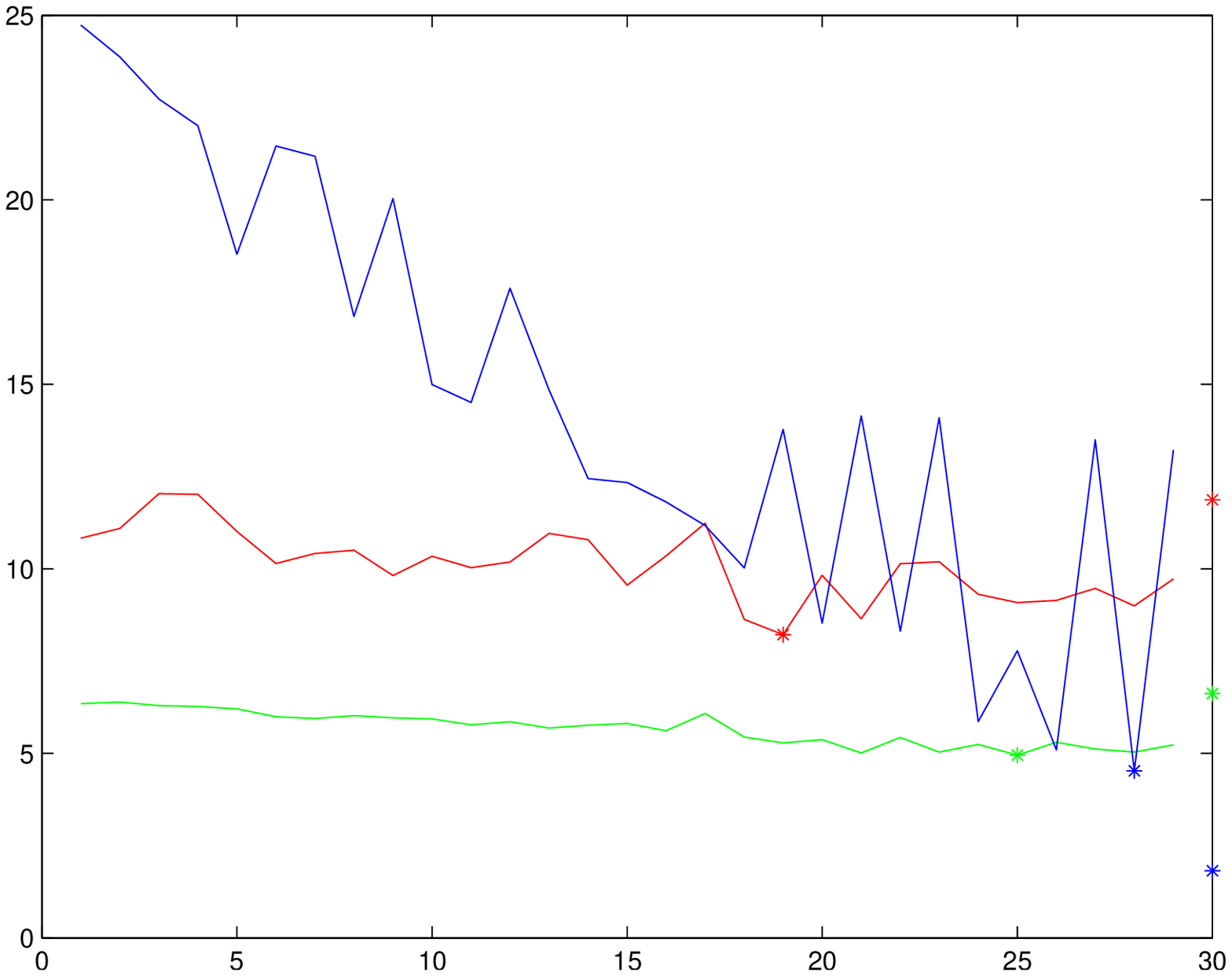,width=2.5 in} &
\psfig{figure=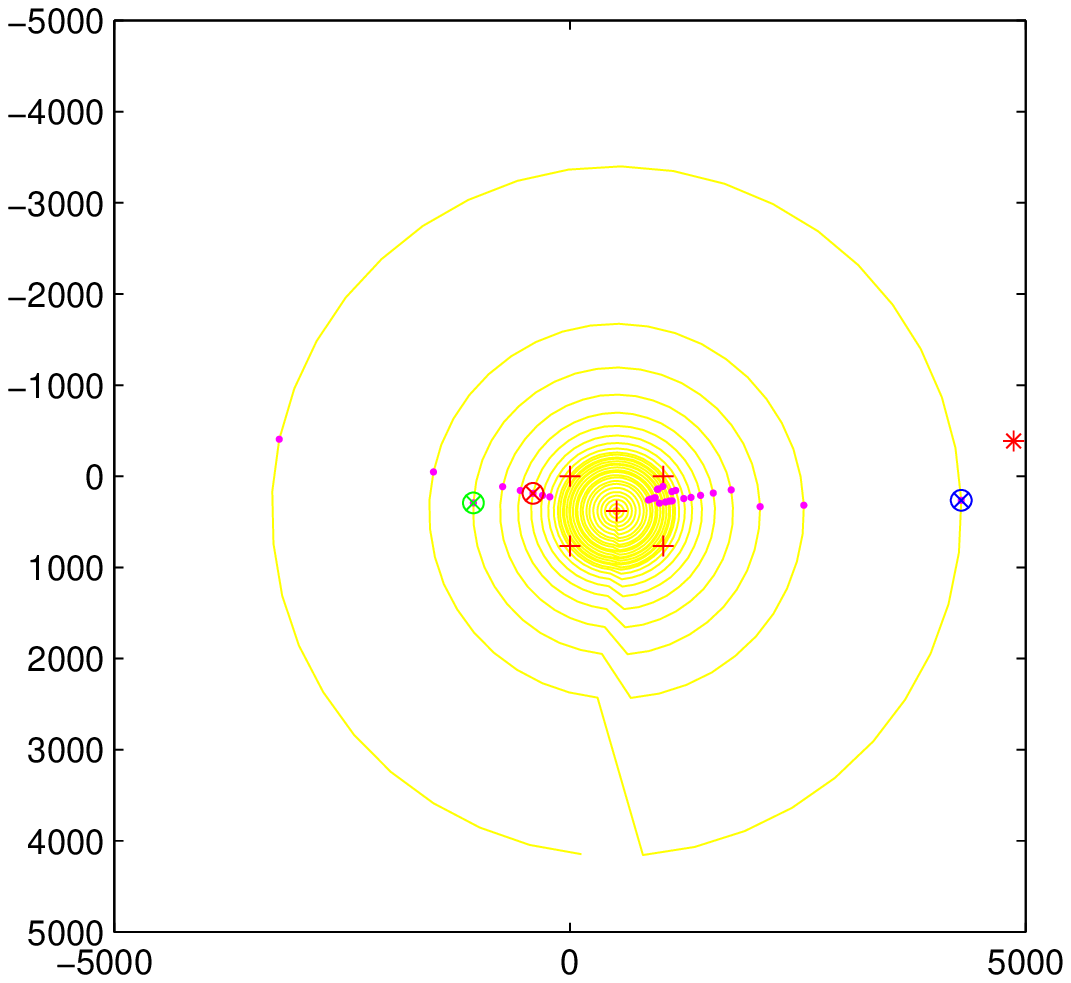,width=2.5 in}\\
(g) & (h)\\
\end{tabular}
\end{tabular}
}
 \caption
{
Case7
 } \label{fig:fige7}
\end{figure}
\begin{figure}[htb]
\centerline{
\begin{tabular}{c}
\psfig{figure=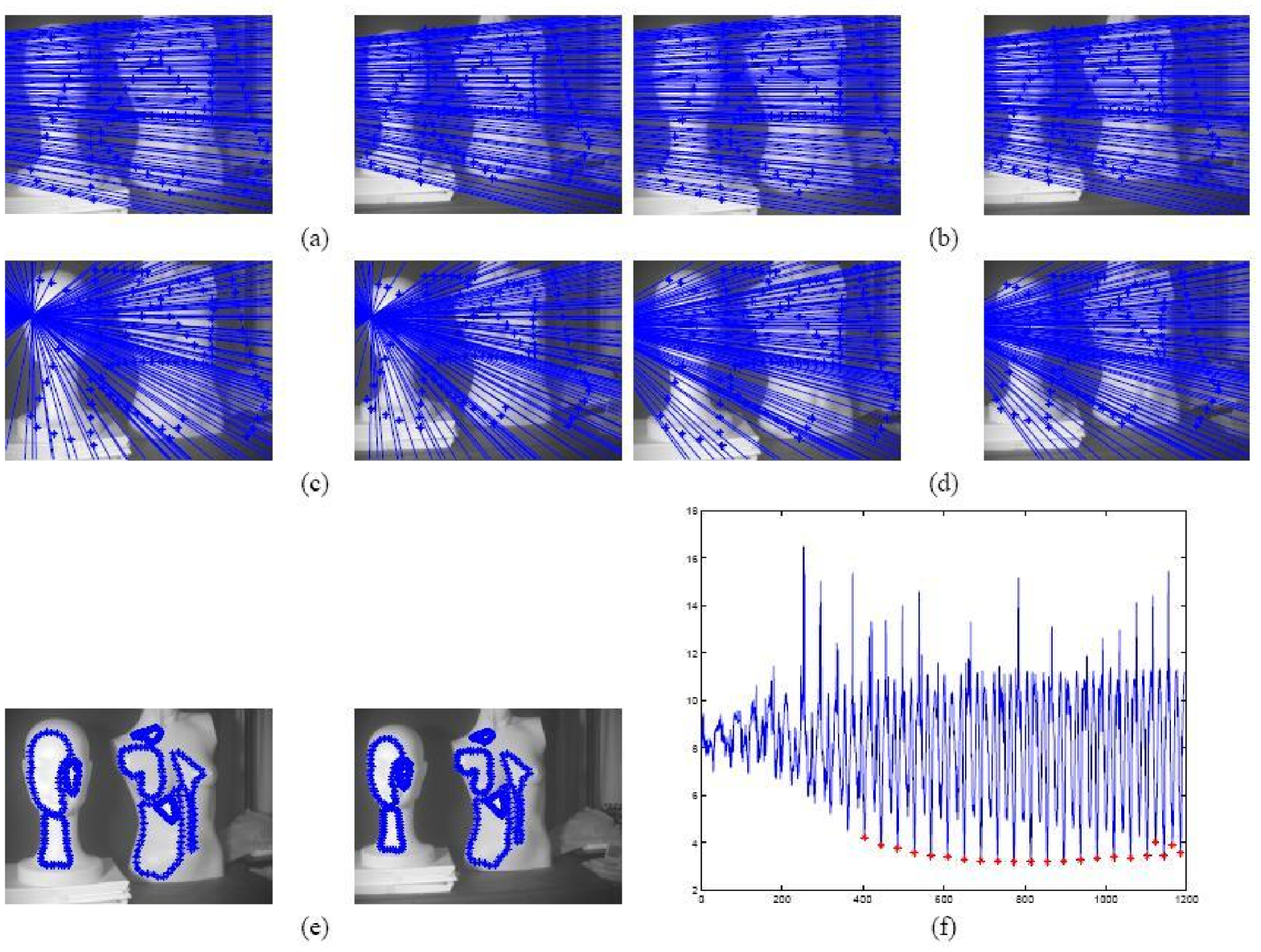,width=5 in}\\
\begin{tabular}{cc}
\psfig{figure=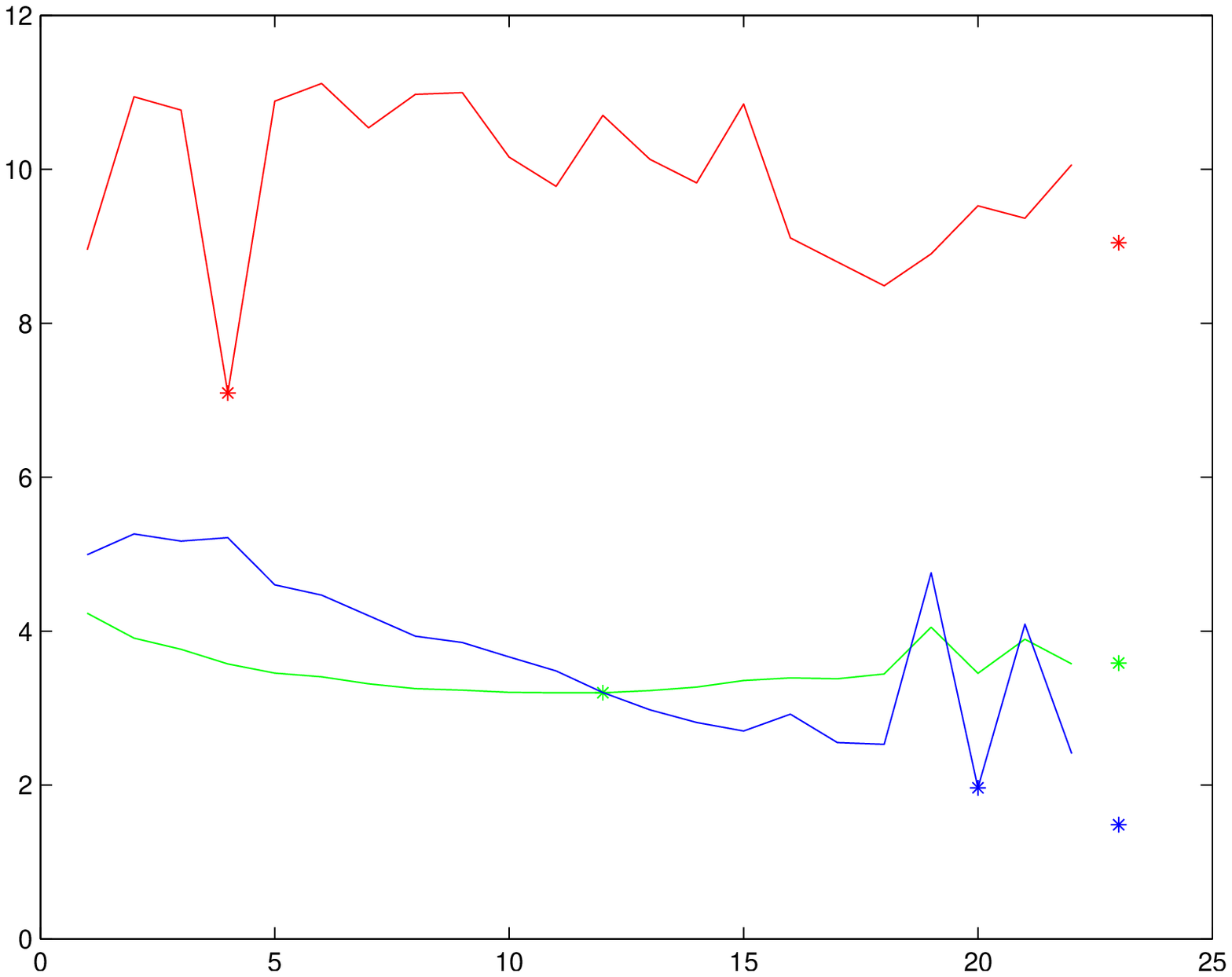,width=2.5 in} &
\psfig{figure=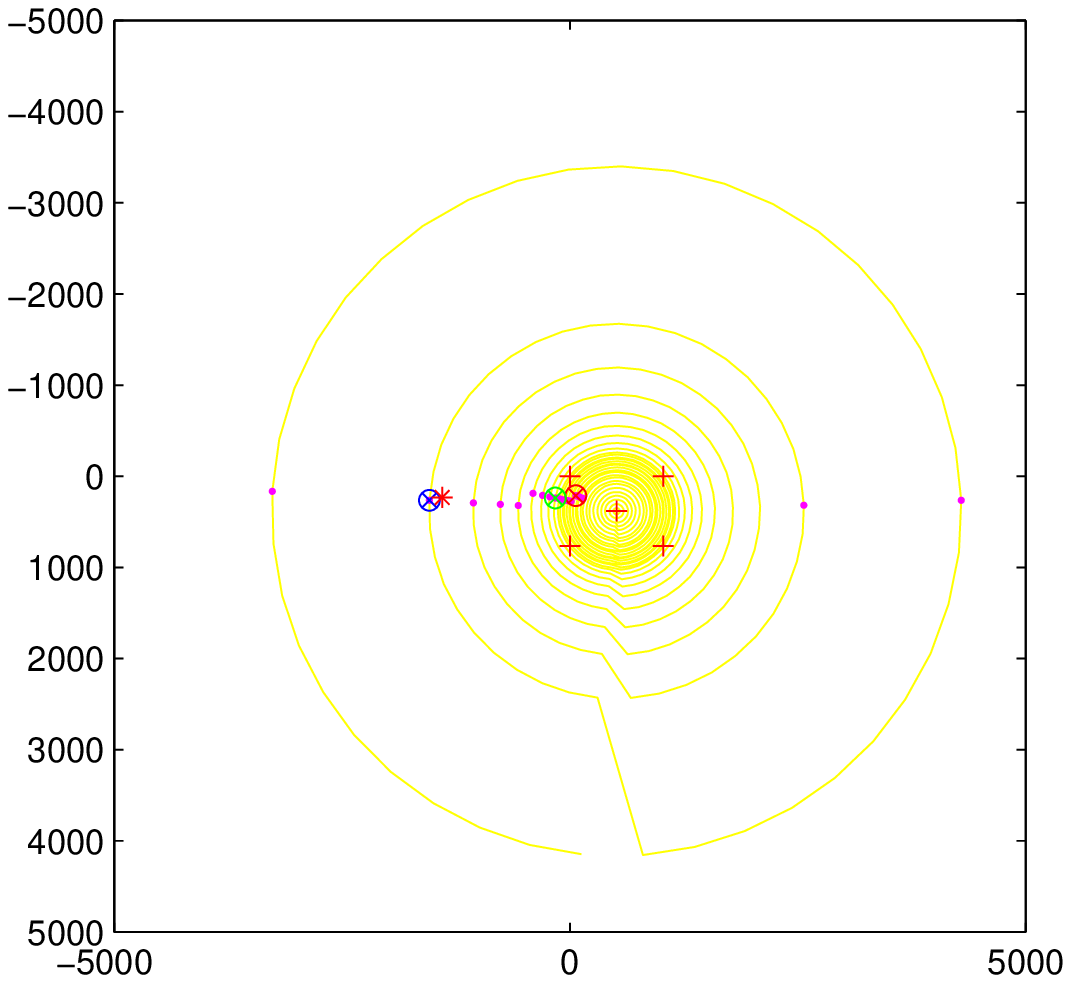,width=2.5 in}\\
(g) & (h)\\
\end{tabular}
\end{tabular}
}
 \caption
{
Case8
 } \label{fig:fige8}
\end{figure}
\begin{figure}[htb]
\centerline{
\begin{tabular}{c}
\psfig{figure=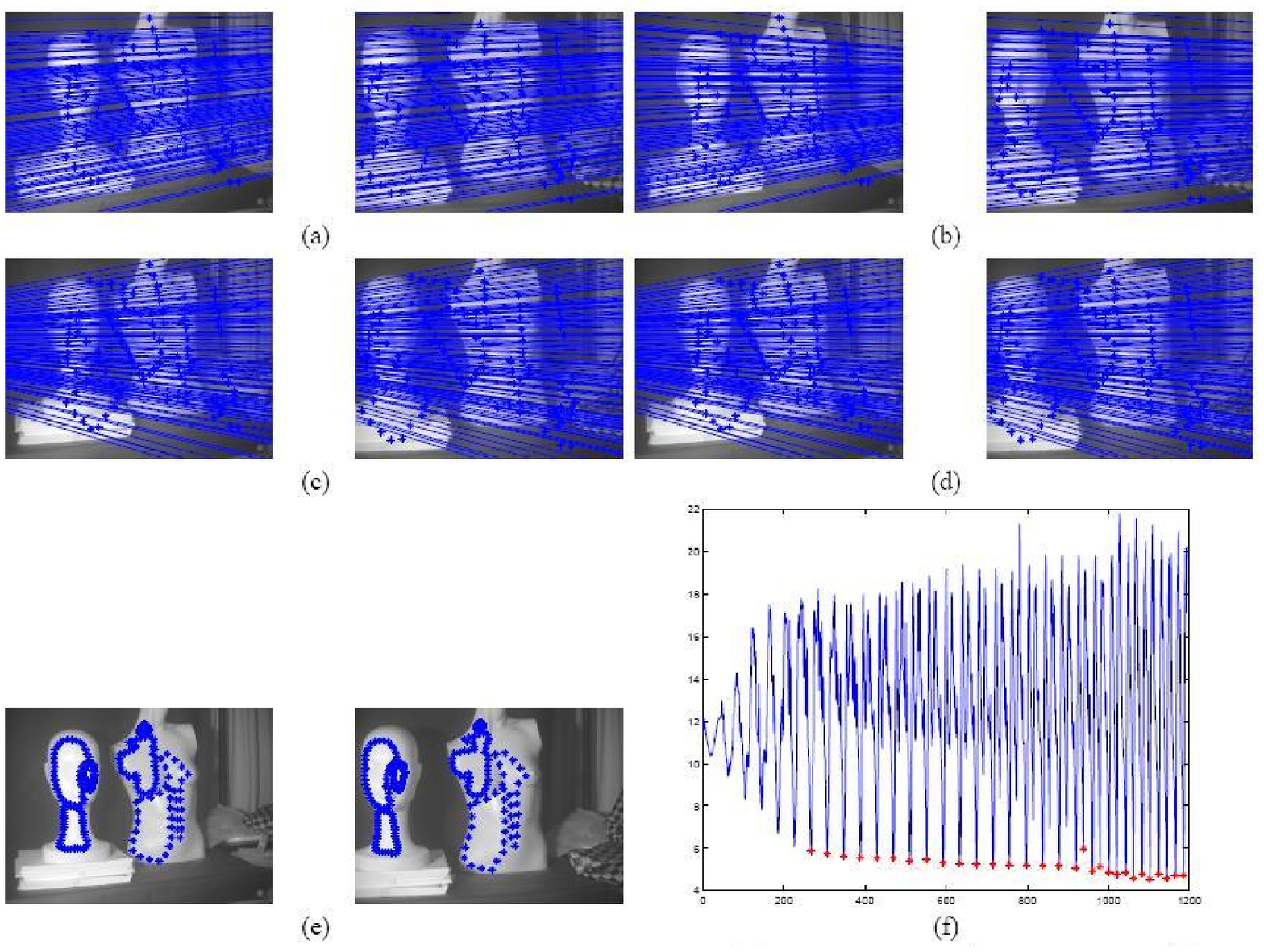,width=5 in}\\
\begin{tabular}{cc}
\psfig{figure=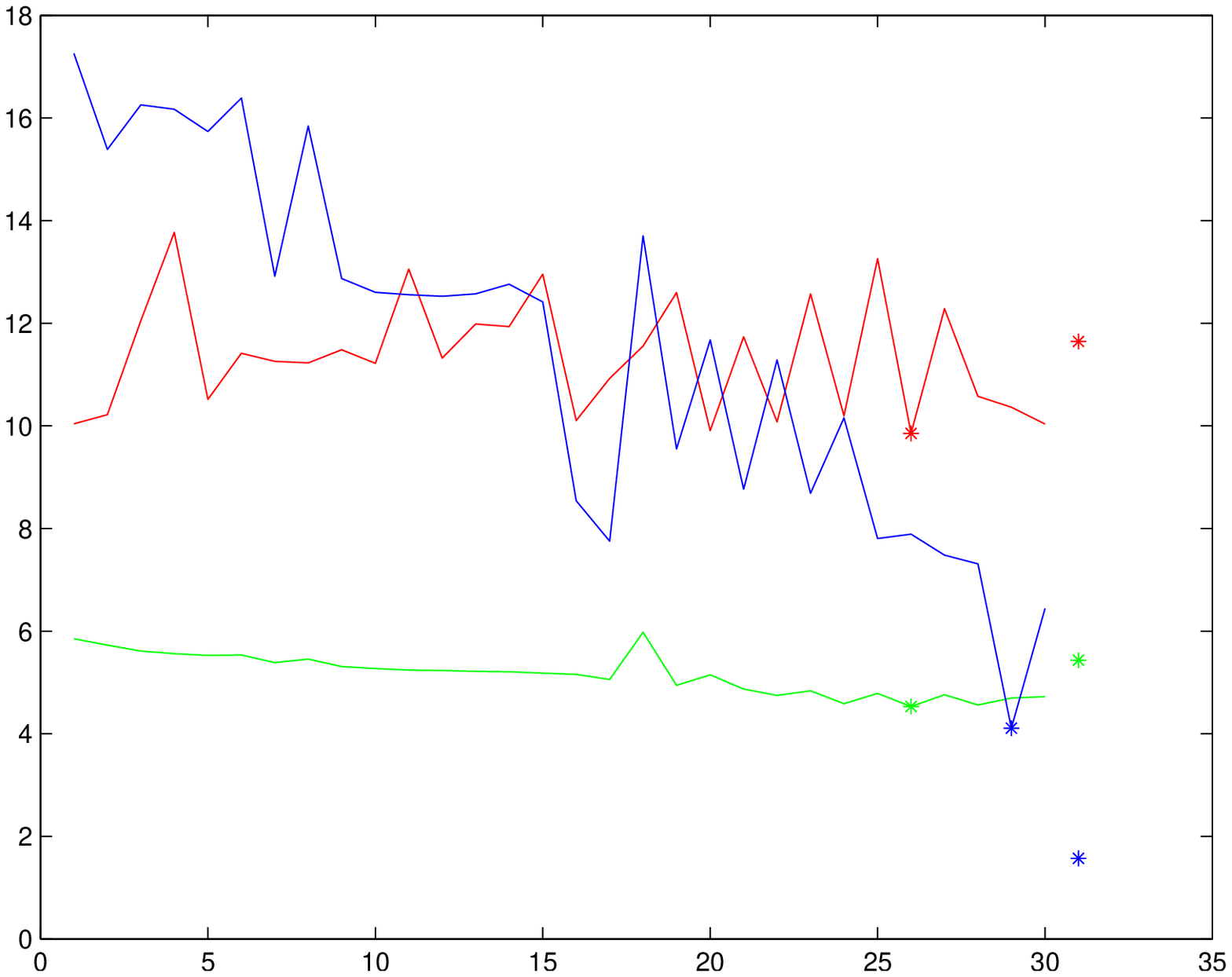,width=2.5 in} &
\psfig{figure=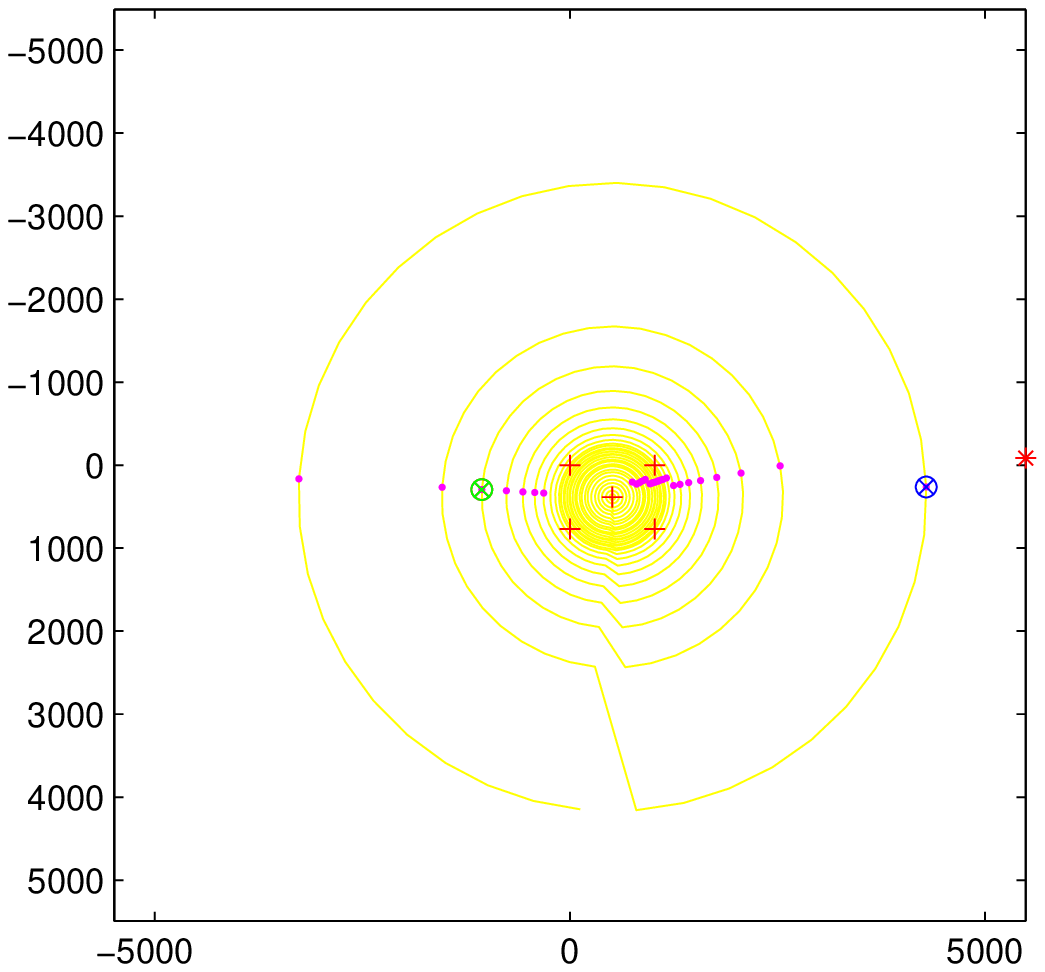,width=2.5 in}\\
(g) & (h)\\
\end{tabular}
\end{tabular}
}
 \caption
{
Case9
 } \label{fig:fige9}
\end{figure}

\end{document}